\pdfoutput=1
\documentclass[11pt]{article}

\usepackage{acl}

\usepackage{times}
\usepackage{latexsym}

\usepackage[T1]{fontenc}
\usepackage[utf8]{inputenc}

\usepackage{microtype}

\usepackage{inconsolata}

\usepackage{graphicx}
\usepackage{algorithm}
\usepackage{algorithmic}
\usepackage{xcolor}
\usepackage{arydshln}
\usepackage{amssymb}
\usepackage{amsmath}
\usepackage{booktabs}
\usepackage{dsfont}
\usepackage{multirow}
\usepackage{tikz}
\usepackage{kotex}

\usepackage{colortbl}
\usepackage{pifont}
\usepackage{tikz}
\usepackage{supertabular}
\usepackage{array}
\usepackage{tabularray}
\usepackage{ragged2e} %
\usepackage{enumitem}
\usepackage{longtable} %

\newcommand{\correct}{\textcolor{green}{\ding{51}}}%
\newcommand{\incorrect}{\textcolor{red}{\ding{55}}}%

\title{Debate Only When Necessary: Adaptive Multiagent Collaboration for Efficient LLM Reasoning}

\author{Sugyeong Eo, Hyeonseok Moon, Evelyn Hayoon Zi, Chanjun Park, Heuiseok Lim\thanks{Corresponding Author} \\
  Korea University \\
  \texttt{\{djtnrud,glee889,evehy,bcj1210,limhseok\}@korea.ac.kr}}
\begin{document}
\maketitle
\begin{abstract}
Multiagent collaboration has emerged as a promising framework for enhancing the reasoning capabilities of large language models (LLMs). Despite improvements in reasoning, the approach introduces substantial computational overhead resulting from iterative agent interactions. Furthermore, engaging in unnecessary debates increases the risk of generating erroneous responses.
To address these challenges, we propose Debate Only When Necessary (DOWN), an adaptive multiagent debate framework that selectively activates debate based on the confidence score of the agent's initial response. Debate is activated only for queries requiring further deliberation, during which agents refine their outputs by referencing peer responses and associated confidence scores.
Evaluations on benchmarks show that DOWN improves efficiency by up to six times while preserving or even outperforming the performance of existing methods. Further analysis indicates that DOWN effectively mitigates the risk of error propagation stemming from the unnecessary debate process. These findings demonstrate the effectiveness of our approach in delivering high-performance LLM solutions at a lower computational cost.
\end{abstract}

\section{Introduction}
Building on the remarkable advancements in large language models (LLMs), recent research has increasingly focused on extending their capabilities to address complex real-world problems~\cite{yao2023react,fan2024bibliometric,chen2024agentverse}. Among various research directions, multiagent collaboration has emerged as a promising approach, inspired by human decision-making processes in complex problem-solving~\cite{minsky1988society,li-etal-2023-theory,chen-etal-2024-reconcile,wang2025mixtureofagents,wu2023autogen,pmlr-v235-du24e}. By engaging in structured debate, LLM agents systematically exchange perspectives and iteratively cross-examine each other's reasoning to refine their responses. This collaborative process facilitates divergent thinking and enhances the reasoning capabilities of LLMs~\cite{liang-etal-2024-encouraging,chen2024agentverse,chan2024chateval}.

\begin{figure}
\centering
\includegraphics[width=
\linewidth]{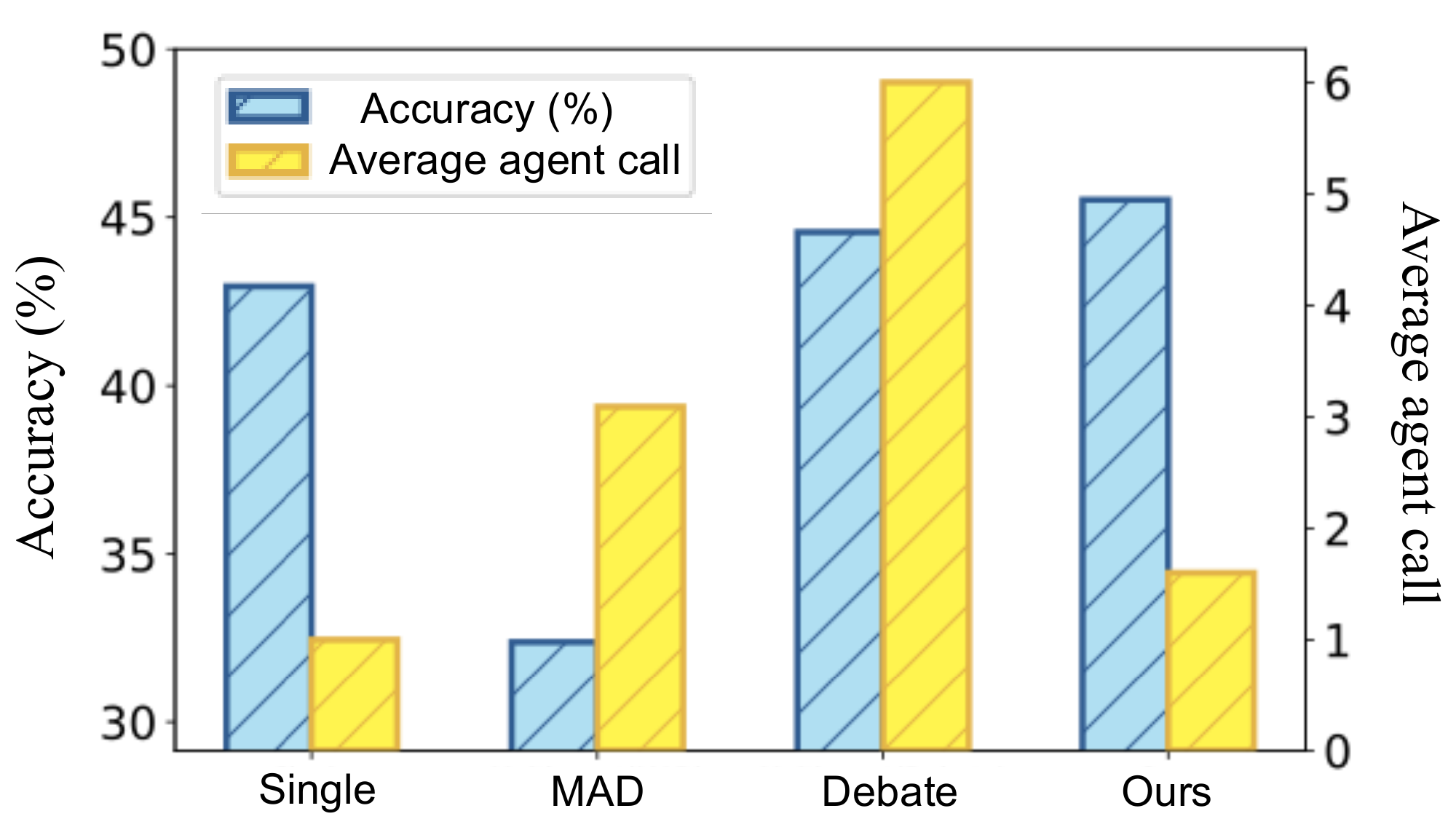}

 \caption{Comparison of accuracy and average agent calls across various multiagent debate methods} \label{fig:acc_eff} 
\end{figure}

Despite these advantages, multiagent collaboration systems exhibit several key limitations. From an efficiency perspective, iterative interactions among agents inherently require multiple agent calls, leading to increased latency and higher inference costs~\cite{snell2024scaling,kapoor2024ai}. Meanwhile, when agents engage in redundant or unnecessary debate, there is a higher likelihood of generating errors that may subsequently be propagated by other agents in the system~\cite{wang-etal-2024-rethinking-bounds}.
Figure~\ref{fig:acc_eff} illustrates both of these issues by plotting the accuracy and average agent calls of different multiagent debate methods. In the Debate system, additional debate rounds improve performance while incurring a sixfold increase in computational overhead. In the MAD framework, despite using more agent calls than the single agent baseline, its accuracy paradoxically declines. Regarding the practical application of multiagent collaboration systems, these challenges emphasize the need for an optimized collaboration approach~\cite{kapoor2024ai,tran2025multi}.

To address these limitations, we propose Debate Only When Necessary (DOWN), an adaptive multiagent collaboration framework that identifies queries requiring debate and selectively facilitates the debate process accordingly.
This framework employs the confidence score to quantify the internal certainty of LLM outputs, a measure that has been extensively utilized to enhance model performance and trustworthiness~\cite{razghandi2025cer, taubenfeld2025confidence, chen-etal-2024-reconcile}.
In the DOWN framework, the initial agent response is generated as the first step. A confidence score is computed during response generation and used to determine whether debate should be initiated.
If the confidence score exceeds a threshold, the debate is skipped, and we regard the initial response as a final answer. If further refinement is required, agents engage in debate to enhance response accuracy. When the collaboration begins, a confidence-guided debate is conducted, during which agents refine their responses by considering both the responses of other participating agents and their confidence scores.
This approach encourages the refinement of responses by utilizing the most persuasive aspects of agent responses.

Our experiments employ models of varying scales, including approximately 8B and 70B parameter models, as well as GPT-4o-mini, evaluated on the MUSR~\cite{sprague2024musr} and StrategyQA~\cite{10.1162/tacl_a_00370} benchmarks. The results demonstrate that adaptive debate invoking significantly reduces computational overhead while maintaining or even surpassing the performance of full-debate baselines. 
Notably, we reveal that this conditional debate serves as a safeguard against cascading errors, effectively enhancing the advantages of agent collaboration.
Our contribution is threefold:

\begin{itemize}
    \item We propose Debate Only When Necessary (DOWN), an adaptive multiagent framework that selectively initiates debate based on the initial response. To the best of our knowledge, this is the first study to explore conditional debate in multiagent systems grounded in initial responses, maximizing efficiency while preserving the benefits of discussion.

    \item Extensive experiments across diverse model sizes and configurations establish the effectiveness of the mechanism. We further find that adaptively engaging debate contributes to mitigating error propagation.
    \item  We establish that the confidence-guided debate process enables the selective integration of reliable responses, emphasizing the effectiveness of multiagent collaboration.
\end{itemize}

\section{Related Work}
\paragraph{LLM-based Multiagent Collaboration}
Drawing inspiration from human collaborative problem-solving behavior, multiagent collaboration systems leverage collective intelligence to improve decision-making. Studies have demonstrated that LLM-powered multiagent systems promote divergent thinking~\cite{xiong-etal-2023-examining,liu2024a,liang-etal-2024-encouraging} and improve reasoning capabilities~\cite{li-etal-2023-theory,yin-etal-2023-exchange,zhuge2023mindstorms}. With these advantages, multiagent collaboration is leveraged for diverse NLP applications: mitigating hallucinations~\cite{fang-etal-2025-counterfactual}, aggregating knowledge across multiple specialized LLMs~\cite{wang2025mixtureofagents}, generating novel scientific ideas and insights~\cite{su2024two}, evaluating LLM-generated responses~\cite{chan2024chateval}, and refining datasets for instruction fine-tuning~\cite{li-etal-2024-coevol}. These advancements highlight the growing impact of multiagent collaboration. 

\begin{figure*}
\centering
\includegraphics[width=\linewidth]{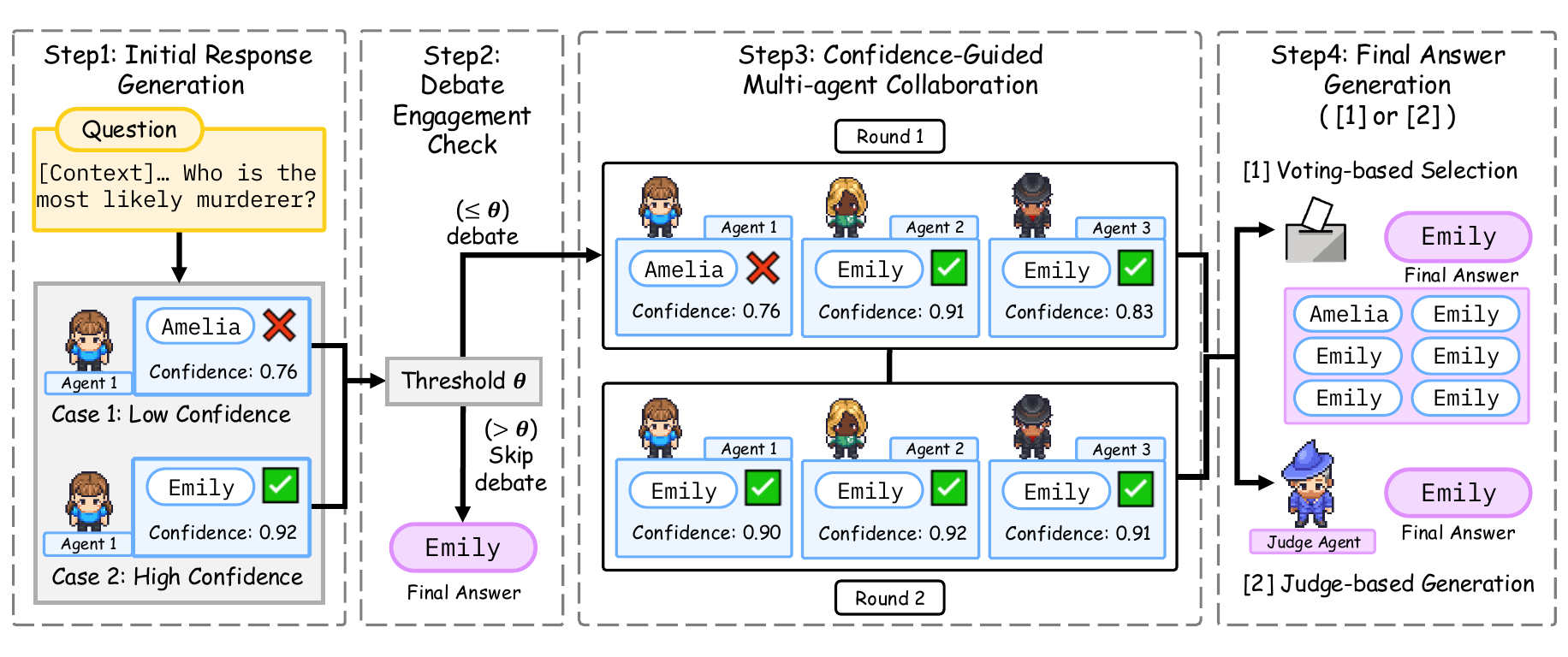}
 \caption{Overview of the Debate Only When Necessary (DOWN) framework. DOWN consists of four stages: (1) the initial agent generates a response, during which the model's confidence score is extracted. (2) if the confidence score exceeds a threshold value, the response is accepted without debate to improve efficiency, otherwise a multiagent debate is activated. (3) agents refine their responses by referencing peer outputs and associated confidence scores. (4) the final answer is selected via majority voting or designated judge agent.} \label{fig:main} 
\end{figure*}

\paragraph{Debate Structures in Multiagent Collaboration Systems}
Recent studies have developed debate structures to optimize the benefits of multiagent collaboration. For instance, \citet{pmlr-v235-du24e} introduces a framework in which agents iteratively refine their responses based on peer-generated outputs. \citet{liang-etal-2024-encouraging} propose a structured debate format that assigns distinct roles to encourage divergent thinking. 
\citet{wang-etal-2024-rethinking-bounds} develop a discussion system in which agents are organized into multiple groups to engage in discussions. 
However, iterative debate frameworks face a fundamental trade-off between efficiency and accuracy~\cite{kapoor2024ai,snell2024scaling}.
To this end, \citet{li-etal-2024-improving-multi} restrict discussions to local neighbors, while \citet{zhou-etal-2025-efficient} introduce a shortcut mechanism that shares similarities with our approach. However, both methods still incur non-trivial agent calls. The former requires neighbor interactions, and the latter relies on self-consistency, with both involving at least five agent calls per sample.

Additionally, iterative debates may propagate errors or introduce a trade-off between diversity and quality~\cite{wang-etal-2024-rethinking-bounds,kapoor2024ai,li2025rethinking}. Our approach focuses on these limitations by proposing a resource-efficient and performance-effective debate framework. 

\section{Debate Only When Necessary}
The collaboration framework consists of four steps: (1) initial response generation, (2) debate engagement check, (3) confidence-guided multiagent collaboration, and (4) final response generation. The confidence score derived from the model during initial response generation determines its subsequent progression. If the score exceeds a predefined threshold, the further debate process is skipped. Otherwise, we proceed with the debate, allowing agents to engage in confidence-guided debate with other agents. We illustrate the overall multiagent collaboration framework in Figure~\ref{fig:main}.

\subsection{System Pipeline}
\paragraph{Step 1: Initial Response Generation}
The model's confidence score obtained during initial response generation determines the progression of the collaboration process. Based on its importance, we adopt two strategies for multiagent configurations: (1) A homogeneous-agent configuration where all collaborating agents use the same model (e.g., Llama) and (2) a mixed-agent configuration where the initial agent is randomly selected for each query (e.g., Llama, Qwen, GPT-4o-mini).

For a given query $q$, an initial agent $\mathcal{A}_1$ from the set of agents $\{\mathcal{A}\}_{i=1}^N$ generates an initial response $r_1$ with the reason for the answer. During this process, a confidence score $c_1$ is extracted to quantify the model's certainty. We first obtain the token logit $L(t_i)$, where $t_i\in r_1$, from the hidden representation of the generated response. These logits are then passed through a softmax function to yield the probability distribution $P(t_i)$ over tokens. To obtain a robust estimate of confidence, we apply normalization by averaging the token probabilities over the generated response. The confidence score $c_1$ is defined as:
\begin{equation}
    c_1=\frac{1}{|r_1|}\sum_{i=1}^{|r_1|}P(t_i), \quad\text{where} P(t_i)=\frac{e^{L(t_i)}}{\sum_j{e^{L(t_j)}}}.
\end{equation} 
 For models that cannot access internal logits, we utilize verbalized confidence as an alternative method.

\paragraph{Step 2: Debate Engagement Check}
Along with the initial response, we extract a confidence score $c_1$ reflecting the model’s certainty in its answer. The confidence score is compared against a threshold score $\theta$: 

\begin{itemize}[nosep, leftmargin=*]
    \setlength{\itemsep}{0pt}
    \setlength{\parskip}{0pt}
    \setlength{\parsep}{0pt}
    \item (1) High confidence ($> \theta$): The response is accepted as the final response without further collaboration, optimizing efficiency by avoiding unnecessary computations. 
    \item (2) Low confidence ($\leq$ $\theta$): The initial response is deemed unreliable, activating the collaboration process to improve response quality.
\end{itemize}

In this context, the threshold is a hyperparameter, where a higher threshold places more emphasis on accuracy, while a lower threshold prioritizes efficiency. 

\paragraph{Step 3: Confidence-Guided Multiagent Collaboration}
The collaboration process involves response generation and refinement. In our experiments, we set up an environment with two rounds, each consisting of three agents.
To conduct multiagent collaboration, in round 1, we obtain responses $r_2$ and $r_3$ from additional agents $\mathcal{A}_2$ and $\mathcal{A}_3$, along with their respective confidence scores $c_2$ and $c_3$.  These confidence scores are explicitly concatenated with each response to convey the agent's certainty level in subsequent rounds. 
        
After all agents generate their responses in the first round, the second round begins. Each agent is given responses from the other two agents, excluding its own. Using this additional context, each agent refines its reasoning and generates an updated response. This allows each agent to leverage confident and compelling arguments in the updated response. 

\paragraph{Step 4: Final Answer Generation}
The responses generated in Step 3 serve as ingredients for deriving the final answer. We explore two distinct strategies for final answer output: voting-based selection and judge-based generation.
The voting-based approach determines the final answer by selecting the most frequent response among all agent-generated outputs. We design this majority voting to enhance robustness by leveraging consensus across multiple perspectives, effectively mitigating individual agent biases.
The judge-based approach introduces an additional judging agent, which generates the final response based on all agent outputs. We induce a judging mechanism to update the final response, prioritizing well-supported and coherent arguments.

\subsection{Threshold Selection Strategy}
To select the optimal confidence threshold $\theta^* \in \Theta$ that jointly maximizes predictive accuracy and inference-time efficiency, we introduce a scoring function that incorporates two soft penalty terms for both objectives.
To support high-performance threshold selection, we define a statistically grounded high-performance region for DOWN, derived by computing the one-sided 95\% Wilson lower bound of the maximum observed accuracy. Threshold candidates whose accuracy lies within this region are considered statistically indistinguishable from the best-performing value, whereas those outside the region are penalized via a soft penalty term.
Let $A_{\theta}=\frac{k}{N}$ and $S_{\theta}=\frac{m}{N}$ denote the accuracy and debate skip rate (i.e. efficiency) for a given threshold value $\theta$ in the candidate set $\Theta$.
Given the highest observed accuracy $A_{\theta}^{\max}$ obtained from $k^{\max}$ correct predictions, the Wilson lower bound $\tilde{A}_\theta^{\max}$ is computed as:
\begin{equation}
    \tilde{A}_{\theta}^{\max} = \frac{2k^{\max} + z^2 - z\sqrt{z^2 + 4k^{\max}\left(1 - \frac{k^{\max}}{N}\right)}}{2(N + z^2)},
\end{equation}
where $z = \Phi^{-1}(0.95) = 1.645$. Thresholds with raw accuracy below $\tilde{A}_{\theta}^{\max}$ are subject to soft penalization. 

In parallel, to ensure a balanced trade-off between computational efficiency and performance improvement through debate, we penalize thresholds whose skip rate $S_{\theta}$ falls outside an interval $[s_{\min}, 1-s_{\min}]$. Here, $s_{\min}$ denotes the proportion symmetrically excluded from the lower and upper bounds of the unit interval $[0,1]$.
The total penalty term is defined as: $P_{\theta}=\max(0, \tilde{A}_{\theta}^{\max}-A_{\theta})+\max(0, s_{\min}-S_{\theta})+\max(0, S_{\theta}-(1-s_{\min}))$.

With both penalty terms defined, we compute the overall utility score for each threshold candidate. To ensure that accuracy and efficiency contribute comparably to the final score, we normalize both metrics via min-max scaling:
\begin{equation}
    \hat{A}_{\theta} = \frac{A_{\theta}-A_{\theta}^{\min}}{A_{\theta}^{\max}-A_{\theta}^{\min}+\epsilon}, \hat{S}_{\theta}=\frac{S_{\theta}-S_{\theta}^{\min}}{S_{\theta}^{\max}-S_{\theta}^{\min}+\epsilon},
\end{equation}
where $\epsilon \ll 1$ is a small constant added for numerical stability. We compute the final score as the sum of the normalized accuracy $\hat{A}_{\theta}$ and efficiency $\hat{S}_{\theta}$, with a penalty term $P_{\theta}$ scaled by a weight factor $\lambda$. The optimal threshold is then determined by maximizing this score:
$\theta^*=\arg\max_{\theta\in \Theta}(\hat{A}_{\theta}+\hat{S}_{\theta}-\lambda\cdot P_{\theta})$.

\begin{table*}[]
\centering
\resizebox{.9\textwidth}{!}{
\begin{tabular}{c|cc|cc|cc|cc|cc}
\toprule[1.5pt]
\multirow{2}{*}{\textbf{Method}} & \multicolumn{2}{c|}{\textbf{Llama-3.1 8B}} & \multicolumn{2}{c|}{\textbf{Ministral 8B}} & \multicolumn{2}{c|}{\textbf{GPT-4o-mini}} & \multicolumn{2}{c|}{\textbf{Llama-3.3 70B}} & \multicolumn{2}{c}{\textbf{Qwen-2.5 72B}} \\ \cline{2-11}
 & Acc. & \multicolumn{1}{c|}{AC} & Acc. & \multicolumn{1}{c|}{AC} & Acc. & \multicolumn{1}{c|}{AC} & Acc. & \multicolumn{1}{c|}{AC} & Acc. & AC \\ \midrule[1.5pt]
Single-CoT & 42.95 & \cellcolor[HTML]{D9D9D9}1.00 & \underline{51.06} & \cellcolor[HTML]{D9D9D9}1.00 & 55.75 & \cellcolor[HTML]{D9D9D9}1.00 & 56.33 & \cellcolor[HTML]{D9D9D9}1.00 & 57.80 & \cellcolor[HTML]{D9D9D9}1.00 \\
Self-refine & 39.46 & \cellcolor[HTML]{D9D9D9}6.00 & 36.90 & \cellcolor[HTML]{D9D9D9}6.00 & 54.29 & \cellcolor[HTML]{D9D9D9}6.00 & 53.67 & \cellcolor[HTML]{D9D9D9}6.00 & 58.47 & \cellcolor[HTML]{D9D9D9}6.00 \\
Self-Consistency & 44.70 & \cellcolor[HTML]{D9D9D9}9.00 & 48.24 & \cellcolor[HTML]{D9D9D9}9.00 & 55.88 & \cellcolor[HTML]{D9D9D9}9.00 & \textbf{58.18} & \cellcolor[HTML]{D9D9D9}9.00 & 58.29 & \cellcolor[HTML]{D9D9D9}9.00 \\ \hline
MAD   & \cellcolor[HTML]{FFFFFF}32.39 & \cellcolor[HTML]{D9D9D9}3.09 & 28.67 & \cellcolor[HTML]{D9D9D9}3.01 & 43.23 & \cellcolor[HTML]{D9D9D9}3.02 & \cellcolor[HTML]{FFFFFF}51.22 & \cellcolor[HTML]{D9D9D9}3.00 & 49.13 & \cellcolor[HTML]{D9D9D9}3.04 \\
Debate  & 44.56 & \cellcolor[HTML]{D9D9D9}6.00 & 48.54 & \cellcolor[HTML]{D9D9D9}6.00 & \underline{57.32} & \cellcolor[HTML]{D9D9D9}6.00 & 57.28 & \cellcolor[HTML]{D9D9D9}6.00 & 58.69 & \cellcolor[HTML]{D9D9D9}6.00 \\ \hline
DOWN-Vote   & \underline{45.51} & \cellcolor[HTML]{D9D9D9}1.50 & \textbf{53.71} & \cellcolor[HTML]{D9D9D9}1.48 & 57.09 & \cellcolor[HTML]{D9D9D9}1.80 & \underline{57.80} & \cellcolor[HTML]{D9D9D9}1.02 & \underline{59.39} & \cellcolor[HTML]{D9D9D9}1.28 \\
DOWN-Judge  & \textbf{45.52} & \cellcolor[HTML]{D9D9D9}1.60 & \textbf{53.71} & \cellcolor[HTML]{D9D9D9}1.57 & \textbf{57.35} & \cellcolor[HTML]{D9D9D9}1.96 & \underline{57.80} & \cellcolor[HTML]{D9D9D9}1.03 & \textbf{59.52} & \cellcolor[HTML]{D9D9D9}1.34
 \\ \bottomrule[1.5pt]
\end{tabular}}\caption{Comparison of accuracy (Acc.) and average agent calls (AC) across single-agent methods, multiagent debate systems, and our proposed approach on the MUSR dataset. Single CoT, Self-refine, and Self-consistency are single model-based approaches, while MAD, Debate, and DOWN are multiagent debate-based systems.}\label{tb:main_musr}
\end{table*}
\begin{table*}[]
\centering
\resizebox{.9\textwidth}{!}{
\begin{tabular}{c|cc|cc|cc|cc|cc}
\toprule[1.5pt]
\multirow{2}{*}{\textbf{Method}} & \multicolumn{2}{c|}{\textbf{Llama-3.1 8B}} & \multicolumn{2}{c|}{\textbf{Ministral 8B}} & \multicolumn{2}{c|}{\textbf{GPT-4o-mini}} & \multicolumn{2}{c|}{\textbf{Llama-3.3 70B}} & \multicolumn{2}{c}{\textbf{Qwen-2.5 72B}} \\ \cline{2-11}
 & Acc. & AC & Acc. & AC& Acc. & AC& Acc. & AC& Acc. & AC\\ \midrule[1.5pt]
Single-CoT  & \underline{70.74} & \cellcolor[HTML]{D9D9D9}1.00 &  67.69 & \cellcolor[HTML]{D9D9D9}1.00 & 78.17 & \cellcolor[HTML]{D9D9D9}1.00 & 80.35 & \cellcolor[HTML]{D9D9D9}1.00 & \underline{78.40} & \cellcolor[HTML]{D9D9D9}1.00 \\
Self-refine & 69.54 & \cellcolor[HTML]{D9D9D9}6.00 & 67.69 & \cellcolor[HTML]{D9D9D9}6.00 & 76.42 & \cellcolor[HTML]{D9D9D9}6.00 & 77.73 & \cellcolor[HTML]{D9D9D9}6.00 & 78.17 & \cellcolor[HTML]{D9D9D9}6.00 \\
Self-Consistency & 68.56 & \cellcolor[HTML]{D9D9D9}9.00 &  \underline{68.12} & \cellcolor[HTML]{D9D9D9}9.00 & 79.48 & \cellcolor[HTML]{D9D9D9}9.00 & \underline{80.79} & \cellcolor[HTML]{D9D9D9}9.00 & 77.29 & \cellcolor[HTML]{D9D9D9}9.00 \\ \hline
MAD   & \cellcolor[HTML]{FFFFFF}44.54 & \cellcolor[HTML]{D9D9D9}4.66 & 57.64 & \cellcolor[HTML]{D9D9D9}3.73 & 70.31 & \cellcolor[HTML]{D9D9D9}3.38 & \cellcolor[HTML]{FFFFFF}79.04 & \cellcolor[HTML]{D9D9D9}3.07 & 73.80 & \cellcolor[HTML]{D9D9D9}3.24 \\
Debate & 70.08 & \cellcolor[HTML]{D9D9D9}6.00 & \textbf{70.74} & \cellcolor[HTML]{D9D9D9}6.00 & 79.04 & \cellcolor[HTML]{D9D9D9}6.00 & 80.35 & \cellcolor[HTML]{D9D9D9}6.00 & \textbf{79.91} & \cellcolor[HTML]{D9D9D9}6.00 \\ \hline
DOWN-Vote  & \textbf{71.18} & \cellcolor[HTML]{D9D9D9} 2.53 &  \underline{68.12} & \cellcolor[HTML]{D9D9D9}3.16 & \textbf{80.79} & \cellcolor[HTML]{D9D9D9}1.92 & \textbf{82.53} & \cellcolor[HTML]{D9D9D9}1.07 & 77.73 & \cellcolor[HTML]{D9D9D9}2.64 \\
DOWN-Judge & 69.87 & \cellcolor[HTML]{D9D9D9}2.83 &  \underline{68.12} & \cellcolor[HTML]{D9D9D9}3.59 & \underline{79.91} & \cellcolor[HTML]{D9D9D9}2.10 & \textbf{82.53} & \cellcolor[HTML]{D9D9D9}1.08 & 77.73 & \cellcolor[HTML]{D9D9D9}2.97 \\ \bottomrule[1.5pt]

\end{tabular}}\caption{Comparison of accuracy (Acc.) and average agent calls (AC) across single-agent methods, multiagent debate systems, and our proposed approach on the StrategyQA dataset}\label{tb:main_sqa}
\end{table*}
\section{Experiments}
\subsection{Experimental Setup}
\paragraph{Models.}
We evaluate two agent configurations: homogeneous and mixed. The homogeneous setup leverages a single model across all agents, experimenting with Llama-3.1 8B (\texttt{meta-llama/Llama-3.1-8B-Instruct}), Ministral 8B (\texttt{mistralai/Ministral-8B-Instruct- 2410}), Qwen-2.5 72B (\texttt{Qwen/Qwen2.5-72B- Instruct}), Llama-3.3 70B (\texttt{meta-llama/ Llama-3.3-70B-Instruct}), and GPT-4o-mini (\texttt{gpt-4o-mini}). The mixed configuration utilizes Qwen-2.5 72B, Llama-3.3 70B, and GPT-4o-mini, with the order of model selection randomized for each query. Each debate round consists of three agents, and we conduct a two-round debate where responses are generated in the first round and refined in the second.

\paragraph{Tasks.}
For evaluation, we utilize benchmarks specifically designed to assess reasoning capabilities.
MUSR~\cite{sprague2024musr} evaluates multi-step soft reasoning over free-text narratives, offering a more complex yet realistic reasoning challenge compared to synthetic benchmarks. Assessments are performed on 756 samples drawn from the murder mysteries, object placements, and team allocation subsets.
StrategyQA~\cite{10.1162/tacl_a_00370} requires implicit reasoning, where inference steps must be strategically derived rather than explicitly provided. Evaluations are conducted on the development set comprising 229 samples. 

\paragraph{Baselines.}\label{sec:baselines}
We compare our approach against multiple established reasoning frameworks, along with a single LLM CoT~\cite{NEURIPS2022_8bb0d291} baseline. (1) \textbf{Self-refine}~\cite{NEURIPS2023_91edff07}: Generating an initial response and iteratively refining its output through self-feedback. We design an environment composed of two rounds, with each round comprising three agents. (2) \textbf{Self-Consistency}~\cite{wang2023selfconsistency}: Sampling diverse reasoning paths and deriving the final answer by selecting the most consistent one through majority voting. We sample nine responses for each query. (3) \textbf{MAD}~\cite{liang-etal-2024-encouraging}: Conducting a debate between two agents with opposing perspectives while a moderator selects the most plausible solution or continues the debate if needed. (4) \textbf{Debate}~\cite{pmlr-v235-du24e}: Engaging agents in a structured debate, iteratively refining their responses by incorporating insights from previous exchanges. We design an environment composed of two rounds, with each round comprising three agents.

\subsection{Implementation Details}
We set the temperature to 0.0 to ensure deterministic response generation and limit the maximum sequence length to 512 tokens. The experimental setup maintains consistent configurations across all models. For the $\theta^*$ selection, we define the candidate set as $\Theta=\{0.7, 0.8, 0.9\}$, starting from a threshold that yields a debate skip rate $S_{\theta}$ close to 1. The symmetric margin $s_{\min}$ and the penalty scaling factor $\lambda$ are set to 0.1 and 15.
For the evaluation metrics, we use accuracy as the metric, while efficiency is assessed based on the average number of agent calls or debate skip rate. The experiments are conducted using four 48GB A6000 GPUs. All prompts used in the experiments are provided in Table~\ref{tb:prompt}.

\begin{figure}
\centering
\includegraphics[width=.9\linewidth]{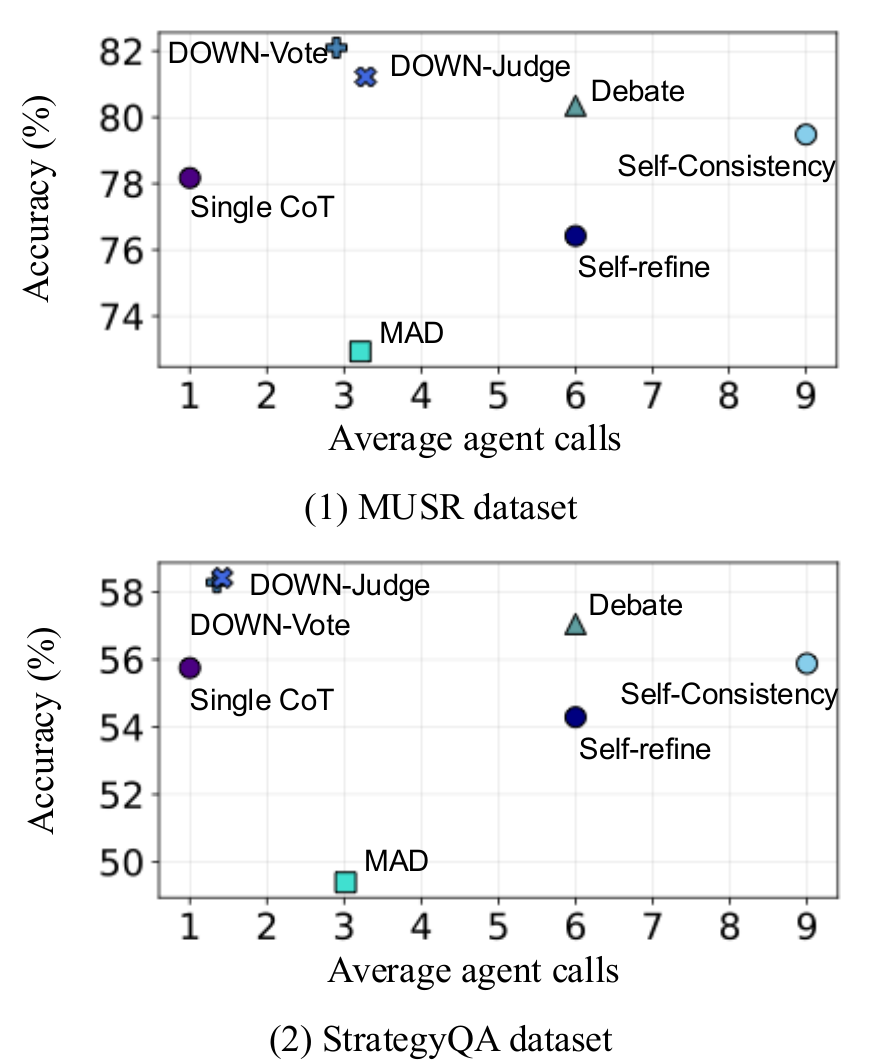} \caption{Comparison of multiagent debate system performance in a mixed-model configuration. The configuration includes Llama3.3-70B, Qwen-2.5 72B, and GPT-4o-mini, with the model order randomized for each query. For single model-based approaches, we present the results of GPT-4o-mini.} \label{fig:mix} 
\end{figure}

\section{Results and Analysis}
\subsection{Main Results}\label{sec:main_results}
\paragraph{Homogeneous-Model Configuration}
Table~\ref{tb:main_musr} presents the results on the MUSR dataset. Unlike prior approaches that require three to nine agent calls per query, our framework achieves comparable or superior performance with only 1.4 calls on average, representing over a sixfold improvement in efficiency. 
The results in Table~\ref{tb:main_sqa} on the StrategyQA dataset further support the effectiveness of our approach.
Compared to Debate and Self-Consistency, our approach achieves comparable or superior performance while significantly reducing the average number of agent calls. Moreover, our framework demonstrates strong robustness across different model families and parameter sizes, confirming its applicability across a wide range of architectures.
These findings highlight that selectively invoking debate, rather than applying it uniformly or relying solely on single agent responses, offers a more principled balance between accuracy and efficiency. 
Interestingly, MAD exhibits lower accuracy compared to other debate-based methods. Consistent with findings from \citet{wang-etal-2024-rethinking-bounds}, we attribute this to its inherent tendency toward contradictory reasoning. While constructive disagreement encourages divergent insights, it also intensifies erroneous reasoning, deteriorating the quality of final responses.
\begin{table}[]
\centering
\resizebox{0.5\textwidth}{!}{
\resizebox{\textwidth}{!}{%
\begin{tabular}{ccccc}
\toprule[1.5pt]
\multicolumn{1}{l}{\textbf{LLM Agent}} & \multicolumn{1}{l}{\textbf{Shift}} & \textbf{MAD} & \textbf{Debate}  & \textbf{Ours} \\ \midrule[1.5pt]
\multirow{2}{*}{GPT-4o-mini} & \correct $\rightarrow$ \incorrect & 70.59 & 50.00 & 33.59 \\
 & \incorrect $\rightarrow$ \correct & 29.41 & 50.00 & 66.41 \\ \hline
\multirow{2}{*}{LLaMA-3.3 70B} & \correct $\rightarrow$ \incorrect & 48.91 & 60.09 & 12.57 \\
 &\incorrect $\rightarrow$ \correct & 51.09 & 39.91 & 87.43 \\ \hline
\multirow{2}{*}{Qwen-2.5 72B} & \correct $\rightarrow$ \incorrect & 63.87 & 50.00 & 39.91 \\
 &\incorrect $\rightarrow$ \correct & 36.13 & 50.00 & 60.09 \\ \hline
\multirow{2}{*}{Mix} & \correct $\rightarrow$ \incorrect & 70.85 & 30.81 & 47.35 \\
 &\incorrect $\rightarrow$ \correct & 29.15 & 69.19 & 52.65 \\ \bottomrule[1.5pt]
\end{tabular}%
}}\caption{Proportions of correct and incorrect response changes before and after debate across multiagent collaboration methods. We denote a correct answer by \correct and an incorrect answer by \incorrect.\label{tb:shift}}
\end{table}
\paragraph{Mixed-Model Configuration}
Under the mixed-model configuration, external feedback originates from models trained on varying datasets and scales, which facilitates the emergence of diverse and complementary reasoning behaviors.
As shown in Figure~\ref{fig:mix},
DOWN consistently strong baselines, achieving higher accuracy with significantly fewer agent calls across both MUSR and StrategyQA. DOWN surpasses the Debate baseline in performance, with a markedly reduced computational burden.
These results demonstrate that our conditional debate remains effective beyond homogeneous setups, adapting seamlessly to mixed-model collaboration. This highlights the robustness and scalability of our debate system in diverse reasoning environments.

\paragraph{Comparison of Final Answer Generation Strategies}
Voting-based selection and judge-based evaluation are two strategies for consolidating multiagent debate responses into a final decision. Experimental results indicate that both approaches yield similar response patterns, with marginal differences depending on the specific setting. However, the judge-based method requires an additional agent call, making it slightly less efficient. Given this trade-off, the voting-based approach is preferable when prioritizing computational efficiency, as it achieves comparable accuracy with fewer computational resources.

% 우리는 debate에서 생성된 responses를 바탕으로 voting-based와 judge-based final answer 도출 방법론을 제안했다. 실험 결과, vote방식과 judge 방식은 상당히 유사한 final response를 내는 경향을 보였다. 두 성능은 적용하는 세팅에 따라 marginal한 차이만을 보인다. 이때 judge 방식의 경우, 한 번의 agent call이 추가되기 때문에 한층 덜 효율적이다. 따라서 효율성 제고를 위해서는 voting 방식을 적용하는 것이 조금 더 추천되는 바이다.
\begin{table}[]\resizebox{.5\textwidth}{!}{
\begin{tabular}{lccc|c}
\toprule[1.5pt]
 \textbf{Model} & \multicolumn{1}{c}{\textbf{Method}} & \begin{tabular}[c]{@{}c@{}}\textbf{Original}\\ \textbf{Debate}\end{tabular} & \begin{tabular}[c]{@{}c@{}}\textbf{Conditional}\\ \textbf{Debate}\end{tabular} & \begin{tabular}[c]{@{}c@{}}\textbf{Skip}\\\textbf{Rate}\end{tabular} \\ \hline 
\multirow{2}{*}{Llama-3.3 70B} & MAD & 79.04 & \textbf{79.91} & 59.83\% \\
& Debate & 80.35 & \textbf{83.41} & 68.56\% \\
 
\multirow{2}{*}{Qwen-2.5 72B} & MAD & 73.80 & \textbf{76.86} & 51.53\% \\ 
 & Debate & \textbf{79.91} & 79.48 & 45.85\% \\ \bottomrule[1.5pt]
\end{tabular}}\caption{Results of applying conditional debate to existing debate systems on the StrategyQA dataset\label{tb:conditional_debate}}
\end{table}
\subsection{Response Shifts in Accuracy}
Table~\ref{tb:shift} reports the proportions of correct and incorrect changes in answers, computed over StrategyQA samples where the final answer differs from the initial prediction. To deepen our investigation, we set the confidence threshold to 0.9, increasing the number of cases where debate is triggered.
Across all model configurations, DOWN consistently achieves a higher successful correction rate than baseline debate systems. Notably, for the LLaMA-3.3-70B model, 87.43\% of changed answers reflect successful corrections of initially incorrect predictions. In contrast, MAD shows a high rate of incorrect changes, frequently revising correct initial answers into incorrect ones, consistent with the trends observed in \S~\ref{sec:main_results}.
The Debate baseline shows a similar rate of correct and incorrect changes, indicating a limited capacity to prioritize reliable peer inputs. In contrast, the high success rate of corrections achieved by DOWN stems from two core design principles. It initiates debate only when the model's confidence is low, thereby avoiding unnecessary changes to already correct answers. When debate is triggered, it refers to peer responses with their associated confidence scores, enabling the model to incorporate more reliable inputs.

\subsection{Effects of Conditional Debate in Multiagent Debates}
To assess the effectiveness of conditional debate, we apply it to MAD and Debate frameworks and evaluate performance on the StrategyQA dataset using models with approximately 70B parameters.
Our findings in Table~\ref{tb:conditional_debate} reveal that applying conditional debate to existing debate methods mostly improves performance. In particular, the Debate framework achieves about a 3\% point accuracy increase on LLaMA-3.3 70B, while MAD exhibits a similar improvement on Qwen-2.5 72B. Despite these gains, debate skip rates remain high at 59.83\% and 51.53\%, respectively, demonstrating remarkable efficiency improvements. These results demonstrate that conditional debate not only enhances efficiency but also positively impacts overall model performance by selectively engaging discussions at appropriate points.

\begin{table}[]
\centering
\resizebox{.5\textwidth}{!}{%
\begin{tabular}{lcc}
\toprule[1.5pt]
\multicolumn{1}{c}{\textbf{Method}} & \multicolumn{1}{c}{\textbf{Acc.}} & \multicolumn{1}{c}{\textbf{AC}} \\ \midrule[1.5pt]
DOWN ($\theta^*$) & 71.18 & 2.53 \\ \hline\hline
\multicolumn{3}{c}{\textit{Debate}}\\
(1) w/o confidence-based scoring in debate & 68.12 & 2.46 \\
 (2) w/o multi-agent debate (single response only) & 70.74 & 1.00 \\
(3) w/o debate skipping (debate only) & 70.74 & 6.00 \\
\multicolumn{3}{c}{\textit{Threshold}}\\ 
(4) w/ lowered threshold ($\theta^* - 0.1$) & 69.87 & 1.20 \\
(5) w/ raised threshold ($\theta^* + 0.1$) & 71.62 & 3.53 \\\bottomrule[1.5pt]
\end{tabular}
}\caption{\label{tb:ablation_confidence}Ablation study on confidence score using the StrategyQA dataset}
\end{table}
\begin{figure}
\centering
\includegraphics[width=
1\linewidth]{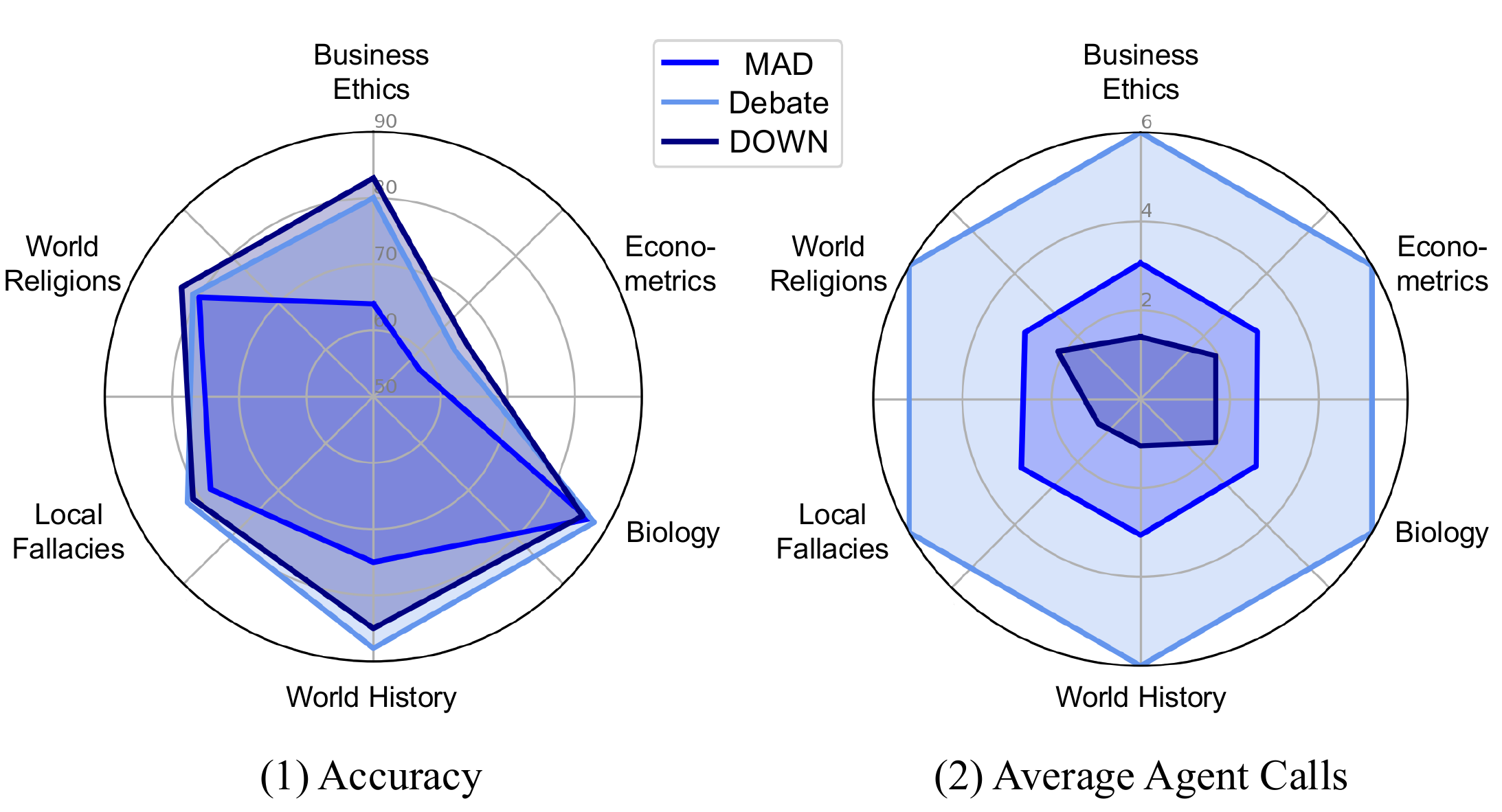}\caption{Accuracy and average agent calls (AC) of multiagent debate methods across six MMLU domains\label{fig:mmlu}}
\end{figure}

\begin{figure*}
\centering
\includegraphics[width=
\linewidth]{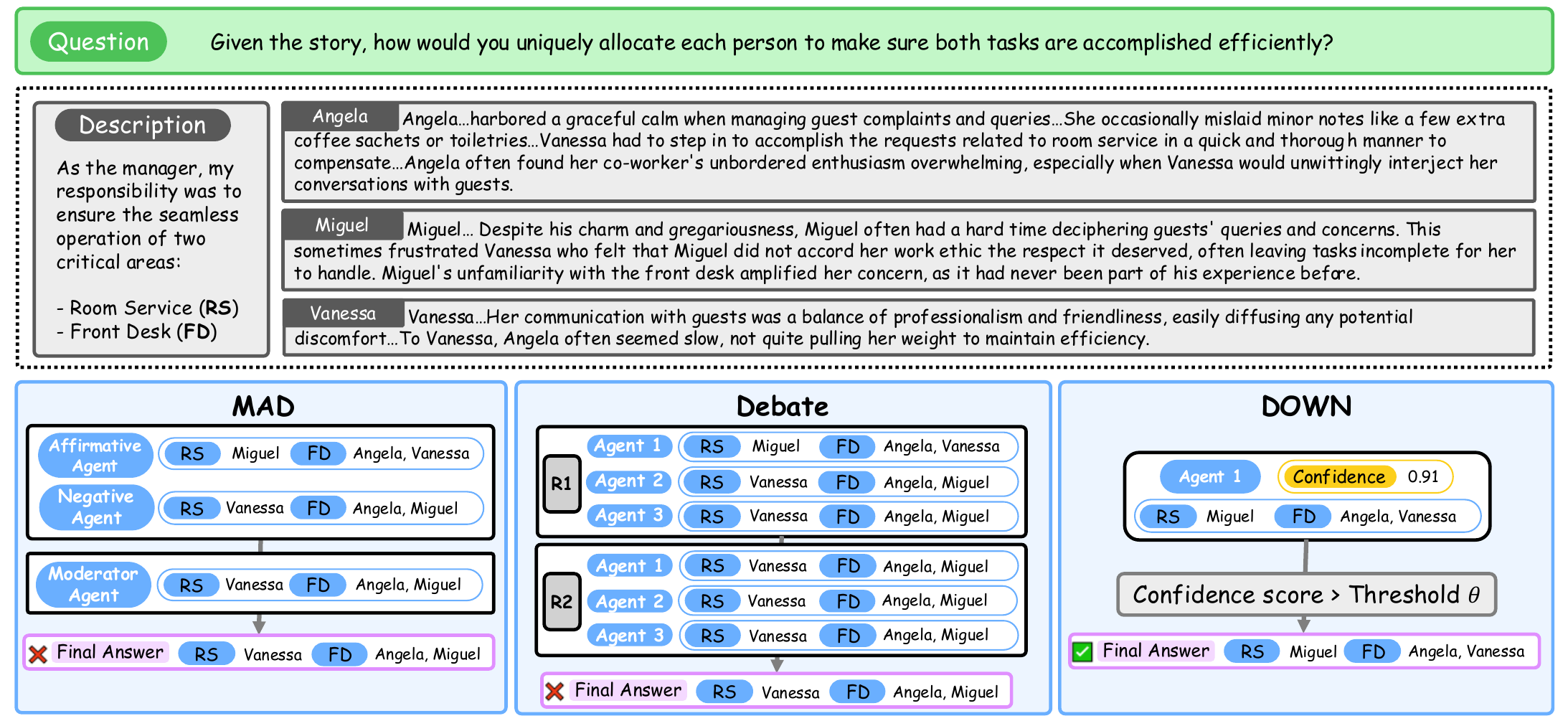}\caption{Qualitative analysis of the MUSR dataset} \label{fig:qualitative_analysis} 
\end{figure*}

\subsection{Ablation Study}
To analyze the contribution of each component in the DOWN framework, we conduct an ablation study on the StrategyQA using the LLaMA-3.1 8B model. The results are shown in Table~\ref{tb:ablation_confidence}.
We examine the role of confidence scores shared among agents during debate. Removing these signals leads to a 3.06\% point drop in accuracy, indicating that confidence serves as an informative cue for selectively incorporating peer responses during the debate process.
When the debate process is either entirely removed or enforced for every input, the accuracy drops to 70.74 in both cases. These findings indicate that always engaging in debate is redundant, yielding no improvement over the single agent baseline.
The next analysis focuses on the threshold selection strategy. To assess its effectiveness, we vary the threshold by $\pm 0.1$ and observe the resulting impact on performance. Modifying the threshold above or below the selected value reveals a trade-off between accuracy and average agent calls. Specifically, the scores obtained for the threshold values $\{\theta^*-0.1, \theta^*, \theta^*+0.1\}$ yields \{0.29, 1.19, 1.00\}, respectively. The chosen threshold $\theta^*$ yields the highest overall score, effectively balancing accuracy and efficiency. It maintains accuracy within a statistically reliable range, as determined by the Wilson lower bound, while substantially reducing computational cost.

\subsection{Generalization analysis on MMLU}
To assess the generalization ability of the DOWN framework across diverse task types and knowledge domains, we additionally evaluate its performance on MMLU using GPT-4o-mini. The performance of different multiagent debate systems across six domains is presented in Figure~\ref{fig:mmlu}. Experimental results show that DOWN achieves performance comparable to that of the Debate method across tasks. Given that the Debate approach consistently requires an average of six agent calls, the DOWN framework shows significantly higher efficiency. The results suggest that DOWN generalizes well, achieving robust performance on benchmarks evaluating both reasoning capabilities and factual knowledge across diverse domains. This highlights its potential to serve as a scalable and effective alternative to full multiagent debate systems.

\subsection{Qualitative Analysis}
Figure~\ref{fig:qualitative_analysis} presents a qualitative comparison of responses generated by different debate systems on the MUSR dataset. The results show that existing debate mechanisms introduce unnecessary modifications. Although the initial responses produced by the Debate and MAD methods are correct, subsequent iterative revisions lead to an incorrect final prediction. This implies the potential risk of error propagation when redundant debate occurs.
In contrast, with an initial confidence score of 0.91, DOWN skips the debate and directly adopts the initial response as the final answer. This suggests that selectively engaging in debate under high-confidence conditions prevents unnecessary modifications while maintaining efficiency. To further substantiate these findings, we present additional qualitative analysis in Appendix~\ref{app:qualitative_analysis}.

\section{Conclusion}
This work proposed the DOWN framework to address the computational inefficiencies and error propagation challenges in multiagent collaboration. By conditionally activating debate based on the model confidence score, the approach significantly reduced computational overhead while preserving or even improving performance. The results showed that conditional debate enhanced efficiency and mitigated cascading errors, leading to more stable reasoning behavior. Furthermore, the confidence-guided multiagent debate amplified the influence of reliable responses on final decisions. These findings established DOWN as an effective optimization strategy, offering a high-performance and efficient solution for multiagent collaboration systems.
multiagent collaboration systems.

\section*{Limitations}
While our proposed DOWN framework demonstrates strong efficiency and robustness, several limitations remain.
This study primarily focuses on English debates, which may limit its applicability to multilingual multiagent collaboration settings. Extending our method to multilingual LLMs would provide deeper insights into its robustness across diverse linguistic contexts.
Due to computational constraints, we employ the GPT-4o-mini instead of the GPT-4o model. Applying the GPT-4o model to our system could further deepen the understanding of our DOWN framework.

\bibliography{custom}

\appendix

\section{Prompts leveraged in DOWN framework}
\begin{table*}[]\resizebox{\textwidth}{!}{
\begin{tabular}{l|ll}
\toprule[1.5pt]
\textbf{Purpose} & \textbf{Prompt} \\ \hline
Initial Answer Generation& \begin{tabular}[c]{@{}l@{}}[debate topic] Please output your answer in json format, \\ with the format as follows: \{\textbackslash{}"base\_reason\textbackslash{}": \textbackslash{}"\textbackslash{}", \textbackslash{}"base\_answer\textbackslash{}": \textbackslash{}"\textbackslash{}"\}. \\ Please strictly output in JSON format, do not output irrelevant content.\end{tabular}     &  \\ \hline
Initial Answer Generation (w Confidence) & \begin{tabular}[c]{@{}l@{}}[debate topic] Please output your answer in json format, \\ with the format as follows: \{\textbackslash{}"base\_reason\textbackslash{}": \textbackslash{}"\textbackslash{}", \textbackslash{}"base\_answer\textbackslash{}": \textbackslash{}"\textbackslash{}", \\ \textbackslash{}"confidence\_score\textbackslash{}":range of 0-1\}. Please strictly output in JSON format,\\  do not output irrelevant content.\end{tabular} &  \\ \hline
Answer Update & \begin{tabular}[c]{@{}l@{}}Using the solutions from other agents as additional information, \\ can you provide your answer to the problem?\end{tabular}  
&  \\ \hline
Judge prompt & \begin{tabular}[c]{@{}l@{}}Based on the following responses, generate an updated response that most accurately \\ addresses the given query. Query: [debate topic] Responses: [agent responses]. \\Summarize your reasons for selecting this side and provide the final answer.\end{tabular}  
&  \\
\bottomrule[1.5pt]
\end{tabular}}\caption{\label{tb:prompt}Prompt configuration leveraged in our experiments}
\end{table*}
We present the prompts utilized in our experiments in Table~\ref{tb:prompt}. These prompts are applied throughout the DOWN framework to ensure structured and consistent response generation.

\section{Qualitative Analysis of Debate Systems}\label{app:qualitative_analysis}
Additional qualitative analysis for each multiagent collaboration method is presented in Table~\ref{tb:qual1}, Table~\ref{tb:qual2}, and Table~\ref{tb:qual3}.

Table~\ref{tb:qual1} presents the responses for different debate methods on the StrategyQA dataset. While the initial response in the MAD system is correct, the system revises its answer after being influenced by the opposing stance of the negative debater. While this process encourages divergent thinking, it ultimately leads to incorrect conclusions due to error propagation. In contrast, the Debate methodology consistently generates correct responses across all cases. Although this approach required six agent calls, it enhances response reliability. The DOWN methodology, on the other hand, produces an initial response with a confidence score of 0.95, leading the system to skip the debate process. This result highlights the efficiency gains achieved through conditional debate.

Table~\ref{tb:qual2} shows the results on the MUSR dataset. Experimental results indicate that both the MAD and Debate methodologies fail to conduct correct reasoning. The initial response in DOWN shows a confidence score of 0.89, while subsequent responses report 0.84 and 0.92, respectively. Notably, in the second round, the initial agent revised its response based on higher-confidence outputs from other agents, ultimately arriving at the correct answer. Model confidence scores also generally increase in the second round. This finding underscores the importance of confidence scores as a guiding metric, allowing the system to integrate reliable responses.

Table~\ref{tb:qual3} follows a trend similar to Table~\ref{tb:qual1}. The DOWN system strategically skips the debate process, enhancing efficiency while maintaining answer accuracy.

\clearpage
\onecolumn

\centering
{\begin{longtable}
{p{0.1\textwidth}|p{0.1\textwidth}|>{\raggedright\arraybackslash}p{0.1\textwidth}|
>{\raggedright\arraybackslash}p{0.6\textwidth}}

\toprule[1.5pt] 
\textbf{Method} & \textbf{Agent} & \textbf{Category} & \textbf{Content} \\ 
\midrule[1.5pt]
\endfirsthead

\toprule[1.5pt] 
\textbf{Method} & \textbf{Agent} & \textbf{Category} & \textbf{Content} \\ 
\midrule[1.5pt]
\endhead

%\midrule[1.5pt]
%\multicolumn{4}{r}{\textit{Continued on next page}} \\ 
\midrule[1.5pt]
\endfoot

%\bottomrule[1.5pt]
\endlastfoot

Query & \multicolumn{3}{>{\raggedright\arraybackslash}p{0.8\textwidth}}{Could a hundred thousand lolcats fit on a first generation iPhone?} \\ \hline
Answer & \multicolumn{3}{>{\raggedright\arraybackslash}p{0.8\textwidth}}{Yes} \\ \midrule[1.5pt]

% MAD
\multirow{6}{*}{MAD} 
% MAD - Affirmative Debater
& \multirow{2}{*}{\begin{tabular}[c]{@{}l@{}}Affirmative \\ Debater\end{tabular}}
%  MAD - Affirmative Debater - Reason
& Reason 
& The storage capacity of a first generation iPhone is 4GB, 8GB, or 16GB, and the average size of an image file is approximately 100KB. ... However, ... it's theoretically possible for a hundred thousand lolcats to fit on a first generation iPhone with sufficient storage. \\ \cline{3-4}
%  MAD - Affirmative Debater - Answer
&  & Answer & Yes \\ \cline{2-4}
% MAD - Negative Debater
& \multirow{2}{*}{\begin{tabular}[c]{@{}l@{}}Negative \\ Debater\end{tabular}}
%  MAD - Negative Debater - Reason
& Reason 
& No. The first generation iPhone had a maximum storage capacity of 16GB for the 16GB model, but also consider other factors such as the operating system, apps, and other data that would occupy some of that space. ... Considering these constraints, it's unlikely that a hundred thousand lolcats could fit on a first generation iPhone.
 \\ \cline{3-4}
%  MAD - Negative Debater - Answer
&  & Answer & No \\ \cline{2-4}
% MAD - Moderator
& \multirow{2}{*}{\begin{tabular}[c]{@{}l@{}}Moderator\end{tabular}}
%  MAD - Moderator - Reason
& Reason 
& While the affirmative side presents a theoretical calculation that a hundred thousand lolcats could fit within the storage capacity of a higher-end first generation iPhone, the negative side raises ... %crucial points about the practical limitations imposed 
by the operating system, other data, and the file system. These factors significantly reduce the available space for storing images, making it unlikely that a hundred thousand lolcats could be accommodated.
\\ \cline{3-4}
%  MAD - Moderator - Answer
&  & Answer & No \\ \cline{2-4}
% MAD - Final Answer
& \multicolumn{2}{>{\centering\arraybackslash}p{0.2\textwidth}|}{\textbf{Final Answer}} & No \\  \midrule[1.5pt]

\multirow{12}{*}{Debate} & \multirow{2}{*}{\begin{tabular}[c]{@{}l@{}}Round 1 \\ Agent 1\end{tabular}} & Reason & The first generation iPhone had a storage capacity of 4GB, 8GB, ... %or 16GB, and assuming an average size of 100KB per lolcat image, a hundred thousand lolcats would require approximately 10GB of storage space, which is within the storage capacity of the higher-end models 
but exceeds the lower-end models. ... Assuming a very small size for each lolcat to fit on the screen, it's theoretically possible to display them, but it would be extremely impractical and unlikely. \\ \cline{3-4}
&  & Answer & Yes \\ \cline{2-4}
 & \multirow{2}{*}{\begin{tabular}[c]{@{}l@{}}Round 1 \\ Agent 2\end{tabular}} & Reason & The first generation iPhone has a storage capacity of 4GB, 8GB, or 16GB, and assuming an average size of a lolcat image (around 100KB), we can estimate the maximum number of images that can fit. ... Still, theoretically, a hundred thousand lolcats could fit on a 16GB first-generation iPhone if we only consider storage space for images and nothing else. \\ \cline{3-4}
&  & Answer & Yes \\ \cline{2-4} 
& \multirow{2}{*}{\begin{tabular}[c]{@{}l@{}}Round 1 \\ Agent 3\end{tabular}} & Reason & The first generation iPhone had a storage capacity of 4GB, 8GB, or 16GB, and assuming an average image size of 100KB, we can estimate the maximum number of images that can be stored. ... Nevertheless, from a purely storage-based perspective, it is theoretically possible for a hundred thousand lolcats to fit on a 16GB first generation iPhone. \\ \cline{3-4} 
&  & Answer & Yes \\ \cline{2-4}
& \multirow{2}{*}{\begin{tabular}[c]{@{}l@{}}Round 2 \\ Agent 1\end{tabular}} & Reason & Considering the storage capacity of the first generation iPhone, with models ranging from 4GB to 16GB, ... %and estimating an average size of 100KB per lolcat image, a hundred thousand lolcats 
would require approximately 10GB of storage space. ... Given this interpretation and focusing solely on storage capacity, it is theoretically possible for a hundred thousand lolcats to fit on a 16GB first-generation iPhone, especially when considering only the storage aspect and disregarding the practicality of displaying them. \\ \cline{3-4}
 &  & Answer & Yes \\ \cline{2-4}
 & \multirow{2}{*}{\begin{tabular}[c]{@{}l@{}}Round 2 \\ Agent 2\end{tabular}} & Reason & Considering the storage capacity of the first generation iPhone, with models ranging from 4GB to 16GB, and estimating an average size of 100KB per lolcat image, a hundred thousand lolcats would require approximately 10GB of storage space. ... Therefore, focusing solely on storage, a hundred thousand lolcats could theoretically fit on a 16GB first-generation iPhone, albeit with the understanding that displaying them all at once would be highly impractical. \\ \cline{3-4}
 &  & Answer & Yes \\ \cline{2-4}
& \multirow{2}{*}{\begin{tabular}[c]{@{}l@{}}Round 2 \\ Agent 3\end{tabular}} & Reason & Considering the storage capacity of the first generation iPhone, with models ranging from 4GB to 16GB, and assuming an average size of 100KB per lolcat image, a hundred thousand lolcats would require approximately 10GB of storage space. ... Taking into account the solutions from other agents, which also consider the theoretical possibility of storing a hundred thousand lolcats on a 16GB model, it appears that from a storage perspective, it is indeed possible. \\ \cline{3-4}
&  & Answer & Yes \\ \cline{2-4}
& \multicolumn{2}{>{\centering\arraybackslash}p{0.2\textwidth}|}{\textbf{Final Answer}}
% & & \textbf{Final Answer}
& Yes \\ \midrule[1.5pt]
\multirow{3}{*}{DOWN} & \multirow{2}{*}{\begin{tabular}[c]{@{}l@{}}Round 1 \\ Agent 1\end{tabular}} & Reason & The first generation iPhone had a storage capacity of 4GB, 8GB, or 16GB, and each lolcat image is approximately 100KB in size. A hundred thousand lolcats would require 10GB of storage space, which exceeds the capacity of the 4GB and 8GB models but fits within the 16GB model. \\ \cline{3-4}
 &  & Answer & Yes \\ \cline{3-4}
 &  & Confidence Score & 0.95 \\ \cline{2-4}
& \multicolumn{2}{>{\centering\arraybackslash}p{0.2\textwidth}|}{\textbf{Final Answer}}& Yes
\\\bottomrule[1.5pt]
\caption{\label{tb:qual1}Qualitative analysis of the StrategyQA dataset}
\end{longtable}}
\centering
{\begin{longtable}
{p{0.1\textwidth}|p{0.1\textwidth}|>{\raggedright\arraybackslash}p{0.1\textwidth}|
>{\raggedright\arraybackslash}p{0.6\textwidth}}

\toprule[1.5pt] 
\textbf{Method} & \textbf{Agent} & \textbf{Category} & \textbf{Content} \\ 
\midrule[1.5pt]
\endfirsthead

\toprule[1.5pt] 
\textbf{Method} & \textbf{Agent} & \textbf{Category} & \textbf{Content} \\ 
\midrule[1.5pt]
\endhead

\midrule[1.5pt]
\endfoot

%\bottomrule[1.5pt]
\endlastfoot

Query & \multicolumn{3}{>{\raggedright\arraybackslash}p{0.8\textwidth}}{Given the story, how would you uniquely allocate each person to make sure both tasks are accomplished efficiently?\newline\newline  Choices: `Seeking Sponsors and Donations: Amelia, Organizing Event Details: George and Lily', `Seeking Sponsors and Donations: George, Organizing Event Details: Amelia and Lily', `Seeking Sponsors and Donations: Lily, Organizing Event Details: Amelia and George'\newline\newline As the clock ticked closer to our most significant fundraising event, three key players - George, Amelia, and Lily - paced around the office, eager to know their assignments. As their manager, I held the responsibility of delegating two critical tasks among them: the challenging pursuit of Seeking Sponsors and Donations, and the intricate duty of Organizing Event Details.
\newline\newline Amelia, our vibrant member, often found herself with Lily, brainstorming creative ideas in the cafeteria or around the picnic table outside. ... %Amelia spent a considerable part of her life attending charity galas, where she had made a myriad of useful connections. 
She was infamous for her love for detailed schedules and checklists, which often found her volunteering to plan her family reunions that ran smoothly under her supervision. ... %She had recently tasted the sweet fruit of her hard work by raising a significant amount of money for a charity event at her workplace, a fact she wore as a badge of honour.
\newline\newline However, the presence of George often marred her high spirits. George, indeed, was a hard one to deal with. Even though he had this endless passion ... %for fundraising and often initiated conversations with potential sponsors in his free time
 , he often disagreed with Lily's ideas during meetings, publicly criticized Amelia's work enough to dampen her spirit, and worst of all, forgot crucial tasks that he was responsible for.
 ... %Having worked with a marketing firm previously dealing with sponsorships, one would think he would be better at managing his responsibilities. 
 But he often confessed feeling overwhelmed managing multiple tasks at once.
 \newline\newline Then there was Lily, a networking maven. ... %She had connections with a broad spectrum of wealthy friends and acquaintances and was certainly not shy to ask people for money due to her past experience working in sales. 
 Yet, her true forte was her meticulousness that reflected in her attention to detail.
 ... %She had previously worked as an event manager for a catering company and was known for planning her own parties and gatherings with immaculate precision. 
 While she tended to dismiss George's inputs during meetings, she showed a great deal of respect to Amelia and considered her suggestions before making a decision.
 \newline\newline  Now, the ball was in my court. With the fundraising event nearing, it was my responsibility to ensure that these three unique individuals were correctly placed to utilize their strengths best. ... %Having studied their characteristics and interaction dynamics, deciphering their assignments became just a little easier. 
 As the manager, I knew that this decision would shape the fate of our event and define our team's dynamic in the foreseeable future.} \\ \hline
Answer & \multicolumn{3}{>{\raggedright\arraybackslash}p{0.8\textwidth}}{Seeking Sponsors and Donations: George, Organizing Event Details: Amelia and Lily}  \\ \midrule[1.5pt]
\multirow{6}{*}{MAD} & \multirow{2}{*}{\begin{tabular}[c]{@{}l@{}}Affirmative \\ Debater\end{tabular}} & Reason & Based on the characteristics and strengths of each individual, Amelia's experience in attending charity galas and her ability to make connections, as well as her attention to detail, make her a strong candidate for Seeking Sponsors and Donations. ... %Meanwhile, Lily's meticulousness and experience as an event manager, combined with George's passion for fundraising, albeit his need for a more structured environment, make them a suitable team for Organizing Event Details. 
George's tendency to disagree with others and forget tasks suggests that he may not be the best fit for a role that requires independent work or a high level of organization, but his passion for fundraising can still be utilized in a team setting. \\ \cline{3-4}
 &  & Answer & Seeking Sponsors and Donations: Amelia, Organizing Event Details: George and Lily \\ \cline{2-4}
 & \multirow{2}{*}{\begin{tabular}[c]{@{}l@{}}Negative \\ Debater\end{tabular}} & Reason & I disagree with your answer. Lily's experience in sales and her ability to ask people for money, combined with her connections with wealthy friends and acquaintances, make her a strong candidate for Seeking Sponsors and Donations. 
 ... %Her past experience working in sales has likely honed her skills in persuasion and negotiation, which are essential for securing sponsors and donations. \newline* Amelia's attention to detail, love for detailed schedules and checklists, and experience in planning family reunions make her a suitable candidate for Organizing Event Details. Her ability to work well with Lily, as evidenced by their brainstorming sessions, also suggests that they could work well together in this role. \newline* George's passion for fundraising is still valuable, but his tendency to disagree with others and forget tasks suggests that he may be better suited to a role where he can work in a team and have his tasks clearly defined. Organizing Event Details, with Amelia's organizational skills and Lily's event management experience, could provide the structured environment that George needs to thrive.\newline\newline Overall, this allocation plays to the strengths of each individual, while also taking into account their weaknesses and work styles.
 It allows Lily to leverage her sales experience and connections to secure sponsors and donations, while Amelia and George work together to organize the event details, with Amelia's organizational skills helping to keep George on track. \\ \cline{3-4}
 &  & Answer & Seeking Sponsors and Donations: Lily, Organizing Event Details: Amelia and George \\ \cline{2-4}
 & \multirow{2}{*}{Moderator} & Reason & The negative side's allocation takes into account the strengths and weaknesses of each individual more effectively. ... %Lily's experience in sales and connections make her a strong candidate for Seeking Sponsors and Donations. Amelia's attention to detail and organizational skills make her a suitable candidate for Organizing Event Details, and she can help keep George on track.
 George's passion for fundraising can still be utilized in a team setting, and his tendency to disagree with others and forget tasks is mitigated by working with Amelia and Lily. \\ \cline{3-4}
 &  & Answer & Seeking Sponsors and Donations: Lily, Organizing Event Details: Amelia and George \\ \cline{2-4}
 & \multicolumn{2}{>{\centering\arraybackslash}p{0.2\textwidth}|}{\textbf{Final Answer}}
 & Seeking Sponsors and Donations: Lily, Organizing Event Details: Amelia and George \\ \midrule[1.5pt]
\multirow{12}{*}{Debate} & \multirow{2}{*}{\begin{tabular}[c]{@{}l@{}}Round 1 \\ Agent 1\end{tabular}} & Reason & The task of Seeking Sponsors and Donations requires someone with strong connections and the ability to ask for money, which Lily possesses due to her past experience in sales and her network of wealthy friends and acquaintances. ... %On the other hand, Organizing Event Details demands attention to detail and the ability to manage multiple tasks, skills that Amelia has demonstrated through her experience in planning family reunions and charity events. George, despite his passion for fundraising, has shown difficulties in managing tasks and often disagrees with others, making him less suitable for these roles. 
However, his experience in a marketing firm could still be valuable in a supporting role or in a different capacity, but for the two main tasks at hand, Amelia and Lily are the better fit. \\ \cline{3-4}
&  & Answer & Seeking Sponsors and Donations: Lily, Organizing Event Details: Amelia and George \\ \cline{2-4}
 & \multirow{2}{*}{\begin{tabular}[c]{@{}l@{}}Round 1 \\ Agent 2\end{tabular}} & Reason & Based on the characteristics and interaction dynamics of the three individuals, Amelia's strengths in seeking sponsors and donations due to her connections and experience, as well as her attention to detail, make her a strong candidate for Seeking Sponsors and Donations. ... %However, her negative interactions with George make it essential to separate them. Lily's experience in sales, event management, and her meticulousness make her suitable for Organizing Event Details. George's passion for fundraising and experience in managing sponsorships are valuable, but his tendency to disagree with others and forget tasks make it crucial to place him in a role where his strengths can be utilized without causing conflicts or compromising the event's success. Therefore, pairing Amelia with Seeking Sponsors and Donations, and George and Lily with Organizing Event Details would not be the best choice. 
 Instead, considering George's potential to disrupt the team's dynamic, it would be more efficient to have him focus on Seeking Sponsors and Donations, where his interactions are more limited, and utilize Amelia and Lily's collaborative and organized approach for Organizing Event Details. \\ \cline{3-4}
 &  & Answer & Seeking Sponsors and Donations: George, Organizing Event Details: Amelia and Lily \\ \cline{2-4}
 & \multirow{2}{*}{\begin{tabular}[c]{@{}l@{}}Round 1 \\ Agent 3\end{tabular}} & Reason & To efficiently allocate tasks, we need to consider the strengths and weaknesses of each person. ... %Amelia has a strong background in attending charity galas, making connections, and planning events. She is also meticulous and detail-oriented. Lily has a strong network of wealthy friends and acquaintances, is not shy to ask for money, and is meticulous in planning events. George has a passion for fundraising, but often disagrees with others, criticizes their work, and forgets crucial tasks. Given these characteristics, it's best to pair Amelia and Lily for Organizing Event Details, as they complement each other's strengths and have a good working relationship. 
 George, despite his flaws, has a passion for fundraising and experience in managing sponsorships, making him a better fit for Seeking Sponsors and Donations, where his skills can be utilized without disrupting the team's dynamics. \\ \cline{3-4}
 &  & Answer & Seeking Sponsors and Donations: George, Organizing Event Details: Amelia and Lily \\ \cline{2-4}
 & \multirow{2}{*}{\begin{tabular}[c]{@{}l@{}}Round 2 \\ Agent 1\end{tabular}} & Reason & Considering the characteristics and interaction dynamics of the three individuals, as well as the solutions provided by other agents, it becomes clear that Amelia and Lily should be paired for Organizing Event Details due to their complementary strengths, meticulousness, and positive working relationship. ... %George, despite his flaws, has a passion for fundraising and experience in managing sponsorships, which makes him a better fit for Seeking Sponsors and Donations. 
 This allocation allows for the efficient utilization of each person's strengths while minimizing potential conflicts and disruptions to the team's dynamics. \\ \cline{3-4}
 &  & Answer & Seeking Sponsors and Donations: George, Organizing Event Details: Amelia and Lily \\ \cline{2-4}
 & \multirow{2}{*}{\begin{tabular}[c]{@{}l@{}}Round 2 \\ Agent 2\end{tabular}} & Reason & After considering the characteristics, strengths, and weaknesses of each individual, as well as the solutions provided by other agents, it becomes clear that the most efficient allocation of tasks would be to utilize Lily's strong network and sales experience for Seeking Sponsors and Donations, and pair Amelia and George for Organizing Event Details. ... %However, given George's tendency to disrupt the team's dynamic and his difficulty in managing multiple tasks, it's crucial to reconsider his role. Amelia's meticulousness and event planning experience make her an ideal candidate for Organizing Event Details. Lily's strengths in seeking sponsors and donations, combined with Amelia's organizational skills, would make them a formidable team. George's passion for fundraising, although valuable, is outweighed by the potential risks of his behavior affecting the team's performance. 
 Therefore, the most suitable allocation would be to have Lily focus on Seeking Sponsors and Donations, where her skills can be maximized without causing disruptions, and have Amelia and George work together on Organizing Event Details, with close supervision to mitigate any potential issues caused by George's behavior. \\ \cline{3-4}
 &  & Answer & Seeking Sponsors and Donations: Lily, Organizing Event Details: Amelia and George \\ \cline{2-4}
 & \multirow{2}{*}{\begin{tabular}[c]{@{}l@{}}Round 2 \\ Agent 3\end{tabular}} & Reason & After considering the characteristics, strengths, and weaknesses of each individual, as well as the solutions provided by other agents, it becomes clear that the most efficient allocation of tasks would be to utilize Amelia and Lily's collaborative and organized approach for Organizing Event Details. ... %Both Amelia and Lily have demonstrated attention to detail, experience in planning events, and a positive working relationship. On the other hand, George's passion for fundraising and experience in managing sponsorships make him a suitable candidate for Seeking Sponsors and Donations. Although George has a tendency to disagree with others and forget tasks, his interactions in this role would be more limited, reducing the potential for conflict. 
 This allocation allows each individual to contribute their strengths to the tasks, minimizing the impact of their weaknesses and ensuring the success of the event. \\ \cline{3-4}
 &  & Answer & Seeking Sponsors and Donations: George, Organizing Event Details: Amelia and Lily \\ \cline{2-4}
& \multicolumn{2}{>{\centering\arraybackslash}p{0.2\textwidth}|}{\textbf{Final Answer}}& Seeking Sponsors and Donations: George, Organizing Event Details: Amelia and Lily \\ \midrule[1.5pt]
\multirow{12}{*}{DOWN} & \multirow{2}{*}{\begin{tabular}[c]{@{}l@{}}Round 1 \\ Agent 1\end{tabular}} & Reason & The decision is based on the unique skills and strengths of each team member. Amelia's experience in attending charity galas and her ability to make connections, as well as her attention to detail, make her a strong candidate for Seeking Sponsors and Donations. ... %However, her past interactions with George have been strained, so it's best to keep them separate. Lily's networking skills and experience in sales make her a great fit for Seeking Sponsors and Donations as well. On the other hand, George's passion for fundraising and his experience in managing sponsorships are valuable assets, but his tendency to disagree with others and forget tasks make it crucial to pair him with someone who can complement his strengths and compensate for his weaknesses. Organizing Event Details requires attention to detail and strong organizational skills, which Amelia and Lily both possess. 
Given their positive working relationship and respect for each other's opinions, pairing Amelia and Lily for Organizing Event Details would be an effective way to utilize their skills and minimize potential conflicts. 
%I am 89.02\% certain of the accuracy and validity of this response based on the provided information and context. 
\\ \cline{3-4}
 &  & Answer & Seeking Sponsors and Donations: Lily, Organizing Event Details: Amelia and George \\ \cline{3-4}
 &  & Confidence Score & 0.89 \\ \cline{2-4}
 & \multirow{2}{*}{\begin{tabular}[c]{@{}l@{}}Round 1 \\ Agent 2\end{tabular}} & Reason & To efficiently accomplish both tasks, it's crucial to allocate individuals based on their strengths and working dynamics. Amelia's experience with charity events, connections, and organizational skills make her ideal for Seeking Sponsors and Donations. ... %Meanwhile, Lily's meticulousness and event management background pair well with George's passion for fundraising, despite his difficulties with task management and interpersonal conflicts. 
 By placing George with Lily, his enthusiasm can be channeled constructively under her detail-oriented supervision, minimizing the impact of his weaknesses. 
 %I am 84.98\% certain of the accuracy and validity of this response based on the provided information and context. 
 \\ \cline{3-4}
 &  & Answer & Seeking Sponsors and Donations: Amelia, Organizing Event Details: George and Lily \\ \cline{3-4}
 &  & Confidence Score & 0.84 \\ \cline{2-4}
 & \multirow{2}{*}{\begin{tabular}[c]{@{}l@{}}Round 1 \\ Agent 3\end{tabular}} & Reason & The decision is based on the unique strengths and weaknesses of each team member. Amelia's experience in attending charity galas and her ability to create detailed schedules make her a strong candidate for Organizing Event Details. ... %Lily's networking skills and attention to detail also make her a good fit for Organizing Event Details. 
 George's passion for fundraising and experience in managing sponsorships make him a good candidate for Seeking Sponsors and Donations, but his tendency to disagree with others and forget tasks make it important to pair him with a task that has fewer collaborative elements. 
 %I am 92.58\% certain of the accuracy and validity of this response based on the provided information and context. 
 \\ \cline{3-4}
 &  & Answer & Seeking Sponsors and Donations: George, Organizing Event Details: Amelia and Lily \\ \cline{3-4}
 &  & Confidence Score & 0.92 \\ \cline{2-4}
 & \multirow{2}{*}{\begin{tabular}[c]{@{}l@{}}Round 2 \\ Agent 1\end{tabular}} & Reason & After considering the strengths, weaknesses, and dynamics of each team member, as well as the solutions provided by other agents, it becomes clear that the most efficient allocation of tasks would leverage the unique skills of each individual while minimizing potential conflicts. ... %Amelia's strengths in organization and detail, combined with her experience in charity events, make her a strong fit for Organizing Event Details. Lily's meticulousness, event management background, and networking skills complement Amelia's abilities, making them a formidable team for Organizing Event Details. George's passion for fundraising and experience in managing sponsorships are valuable assets for Seeking Sponsors and Donations. 
 Despite his challenges with task management and interpersonal conflicts, his strengths can be effectively utilized in a role that plays to his advantages. 
 %I am 91.57\% certain of the accuracy and validity of this response based on the provided information and context. 
 \\ \cline{3-4}
 &  & Answer & Seeking Sponsors and Donations: George, Organizing Event Details: Amelia and Lily \\ \cline{3-4}
 &  & Confidence Score & 0.91 \\ \cline{2-4}
 & \multirow{2}{*}{\begin{tabular}[c]{@{}l@{}}Round 2 \\ Agent 2\end{tabular}} & Reason & After considering the solutions from other agents and re-evaluating the strengths and weaknesses of each team member, it's clear that Amelia's skills in attending charity galas and creating detailed schedules make her a strong fit for Organizing Event Details. ...% However, her past strained interactions with George suggest it's best to keep them separate. Lily's networking skills and attention to detail are valuable for Organizing Event Details as well. George's passion for fundraising and experience in managing sponsorships are assets for Seeking Sponsors and Donations, but his tendency to disagree and forget tasks requires him to be paired with a task that has fewer collaborative elements or with someone who can manage his weaknesses. 
 Considering these points and the fact that Lily has a strong ability to seek sponsors and donations due to her sales experience and network, and Amelia and Lily have a positive working relationship, the most efficient allocation would be to have Amelia and Lily handle Organizing Event Details, utilizing their organizational skills and positive dynamic, and George, with his passion for fundraising, focus on Seeking Sponsors and Donations, where his strengths can be maximized with minimal disruption. 
 %I am 89.75\% certain of the accuracy and validity of this response based on the provided information and context.
 \\ \cline{3-4}
 &  & Answer & Seeking Sponsors and Donations: George, Organizing Event Details: Amelia and Lily \\ \cline{3-4}
 &  & Confidence Score & 0.89 \\ \cline{2-4}
 & \multirow{2}{*}{\begin{tabular}[c]{@{}l@{}}Round 2 \\ Agent 3\end{tabular}} & Reason & After re-evaluating the strengths, weaknesses, and dynamics of the team members, and considering the solutions from other agents, it becomes clear that Amelia's skills in attending charity galas, making connections, and attention to detail are highly valuable for Seeking Sponsors and Donations. ...%Her past success in raising a significant amount of money for a charity event and her ability to work well with Lily, as evidenced by their brainstorming sessions, also support this allocation. Meanwhile, George's passion for fundraising and experience in managing sponsorships, although overshadowed by his tendency to disagree with others and forget tasks, can be constructively channelled under the supervision of Lily, who has event management experience and meticulousness. This pairing allows for George's enthusiasm to be utilized while minimizing the impact of his weaknesses. 
 Therefore, the most efficient allocation is to have Amelia focus on Seeking Sponsors and Donations, where her strengths can be fully leveraged, and to pair George with Lily for Organizing Event Details, where Lily's attention to detail and event management skills can complement George's passion for fundraising and help manage his weaknesses. 
 %I am 90.26\% certain of the accuracy and validity of this response based on the provided information and context.  
 \\ \cline{3-4}
 &  & Answer & Seeking Sponsors and Donations: Amelia, Organizing Event Details: George and Lily \\ \cline{3-4}
 &  & Confidence Score & 0.90 \\ \cline{2-4}
& \multicolumn{2}{>{\centering\arraybackslash}p{0.2\textwidth}|}{\textbf{Final Answer-Vote}}& Seeking Sponsors and Donations: George, Organizing Event Details: Amelia and Lily \\ \cline{1-4}
% & & \textbf{Final Answer-Vote}& Seeking Sponsors and Donations: George, Organizing Event Details: Amelia and Lily \\\cline{2-4}
 %& & \textbf{Final Answer-Judge}
 & \multicolumn{2}{>{\centering\arraybackslash}p{0.2\textwidth}|}{\textbf{Final Answer-Judge}}& Seeking Sponsors and Donations: George, Organizing Event Details: Amelia and Lily. Reason: After carefully evaluating the strengths, weaknesses, and dynamics of each team member, as well as considering the solutions provided by other agents, it is clear that the most efficient allocation of tasks would leverage the unique skills of each individual while minimizing potential conflicts. ... %Amelia's strengths in organization, attention to detail, and experience in charity events make her a strong fit for either Seeking Sponsors and Donations or Organizing Event Details. However, her past strained interactions with George suggest it's best to keep them separate. Lily's networking skills, event management background, and attention to detail are valuable for Organizing Event Details, and her positive working relationship with Amelia makes them a formidable team. George's passion for fundraising and experience in managing sponsorships are assets for Seeking Sponsors and Donations, but his tendency to disagree with others and forget tasks requires him to be paired with a task that has fewer collaborative elements or with someone who can manage his weaknesses. 
 Considering these points, the allocation that best utilizes the strengths of each team member while minimizing disruptions is to have Amelia and Lily handle Organizing Event Details, and George focus on Seeking Sponsors and Donations, where his strengths can be maximized with minimal disruption.
 \\\bottomrule[1.5pt]
 \caption{Qualitative analysis of the StrategyQA dataset}
\label{tb:qual2}
\end{longtable}

%\begin{table}[h]
%    \caption{Qualitative analysis of the StrategyQA dataset}
%    \label{tb:qual2}
%\end{table}
}
\centering
{\begin{longtable}
{p{0.1\textwidth}|p{0.1\textwidth}|>{\raggedright\arraybackslash}p{0.1\textwidth}|
>{\raggedright\arraybackslash}p{0.6\textwidth}}

\toprule[1.5pt] 
\textbf{Method} & \textbf{Agent} & \textbf{Category} & \textbf{Content} \\ 
\midrule[1.5pt]
\endfirsthead

\toprule[1.5pt] 
\textbf{Method} & \textbf{Agent} & \textbf{Category} & \textbf{Content} \\ 
\midrule[1.5pt]
\endhead

\midrule[1.5pt]
\endfoot

%\bottomrule[1.5pt]
\endlastfoot

Query & \multicolumn{3}{>{\raggedright\arraybackslash}p{0.8\textwidth}}{Given the story, how would you uniquely allocate each person to make sure both tasks are accomplished efficiently?\newline\newline Choices:`Room Service: Vanessa, Front Desk: Angela and Miguel', `Room Service: Angela, Front Desk: Miguel and Vanessa', `Room Service: Miguel, Front Desk: Angela and Vanessa' 
\newline\newline In the vibrant heart of the city, a boutique hotel thrived, its pulse maintained by the constant hum of hospitality. As the manager, my responsibility was to ensure the seamless operation of two critical areas - Room Service and Front Desk. The task of assigning these duties fell to me, with the diligent trio of Miguel, Angela, and Vanessa under my supervision.
\newline\newline Angela, with her seven-year-long tenure in various customer service roles, harbored a graceful calm when managing guest complaints and queries. ... %Her tone of respect could soothe any disgruntled guest. 
But Angela was not without her quirks. ... %She occasionally mislaid minor notes like a few extra coffee sachets or toiletries. Her room service abilities received a motley of reviews from the hotel guests.
At times, Vanessa had to step in to accomplish the requests related to room service in a quick and thorough manner to compensate. Angela often found her co-worker's unbordered enthusiasm overwhelming, especially when Vanessa would unwittingly interject her conversations with guests.
\newline\newline Miguel, on the other hand, was the silver lining of good-natured humor and friendliness that added a personal touch to the hotel's room service deliveries. His go-with-the-flow demeanor, however, sometimes came off as disorganized, especially to Angela who appreciated structure. ... %He thrived when allowed to work at his own pace, chaffing under the rigid step-by-step service rules Angela preferred. 
Despite his charm and gregariousness, Miguel often had a hard time deciphering guests' queries and concerns. This sometimes frustrated Vanessa who felt that Miguel did not accord her work ethic the respect it deserved, often leaving tasks incomplete for her to handle. Miguel's unfamiliarity with the front desk amplified her concern, as it had never been part of his experience before. \newline\newline Vanessa had perfected her role at a high-end hotel reception over a span of five meticulous years. ... %Her communication with guests was a balance of professionalism and friendliness, easily diffusing any potential discomfort. 
Vanessa meticulously paid attention to every minor detail while performing room service tasks, sometimes stepping on Angela's toes without realizing it. To Vanessa, Angela often seemed slow, not quite pulling her weight to maintain efficiency. 
\newline\newline With these intricacies at play, as the hotel manager, I had to make conscious decisions assigning my team between Room Service and Front Desk duties. Balancing Angela, Miguel and Vanessa's distinctive skills, patience, and perspectives, the heart of hospitality continued to beat, painting vivid memories in the hotel guest's minds. \newline} \\ \hline
Answer & \multicolumn{3}{>{\raggedright\arraybackslash}p{0.8\textwidth}}{Room Service: Miguel, Front Desk: Angela and Vanessa}\\ \midrule[1.5pt]
\multirow{6}{*}{MAD} & \multirow{2}{*}{\begin{tabular}[c]{@{}l@{}}Affirmative \\ Debater\end{tabular}} & Reason & Considering the strengths and weaknesses of each person, Angela's experience in customer service and her calm demeanor make her a good fit for the Front Desk. ... %Miguel's charm and friendliness are better suited for Room Service, where he can interact with guests in a more relaxed environment. 
Vanessa's attention to detail and professionalism make her a good fit for Room Service as well, where she can ensure that tasks are completed efficiently.
 \\ \cline{3-4}
 &  & Answer & Room Service: Miguel, Front Desk: Angela and Vanessa \\ \cline{2-4}
 & \multirow{2}{*}{\begin{tabular}[c]{@{}l@{}}Negative \\ Debater\end{tabular}} & Reason & I disagree with your answer. My allocation would be:
\newline\newline
Room Service: Vanessa, Front Desk: Angela and Miguel
\newline\newline
Here's my reasoning:
\newline\newline* Vanessa's attention to detail and professionalism make her a perfect fit for Room Service, where she can ensure that tasks are completed efficiently and accurately. ... % Her experience in a high-end hotel reception has also honed her skills in handling guest requests and providing excellent service.
\newline* Angela's experience in customer service and her calm demeanor make her a good fit for the Front Desk, where she can handle guest complaints and queries in a professional and respectful manner. ... % Her ability to soothe disgruntled guests is a valuable asset in this role.
\newline* Miguel's charm and friendliness are also well-suited for the Front Desk, where he can interact with guests and provide a warm welcome. ... % Although he may struggle with deciphering guests' queries and concerns, Angela's experience and calm demeanor can help to support him in this role. Additionally, Miguel's go-with-the-flow demeanor may be less of an issue at the Front Desk, where tasks are often more structured and routine.
\newline\newline Overall, ... %this allocation plays to the strengths of each team member, while also providing a supportive environment for them to work together effectively. 
Vanessa's efficiency and attention to detail in Room Service can help to compensate for Angela's occasional misplacement of minor notes, while Angela and Miguel's combined strengths at the Front Desk can provide a warm and professional welcome for guests. \\ \cline{3-4}
 &  & Answer & Room Service: Vanessa, Front Desk: Angela and Miguel \\ \cline{2-4}
 & \multirow{2}{*}{Moderator} & Reason & The negative side's allocation takes into account the strengths and weaknesses of each team member more effectively. ... %Vanessa's attention to detail and professionalism are well-suited for Room Service, while Angela's experience and calm demeanor make her a good fit for the Front Desk. Miguel's charm and friendliness are also well-suited for the Front Desk, where he can be supported by Angela's experience. 
 This allocation also addresses the potential issues with each team member, such as Angela's occasional misplacement of minor notes and Miguel's difficulty with deciphering guests' queries and concerns.\\ \cline{3-4}
 &  & Answer & Room Service: Vanessa, Front Desk: Angela and Miguel \\ \cline{2-4}
& \multicolumn{2}{>{\centering\arraybackslash}p{0.2\textwidth}|}{\textbf{Final Answer}}
% & & \textbf{Final Answer} 
& Room Service: Vanessa, Front Desk: Angela and Miguel \\ \midrule[1.5pt]
%%%%
\shrinkheight{-2\normalbaselineskip}
\multirow{12}{*}{Debate} & \multirow{2}{*}{\begin{tabular}[c]{@{}l@{}}Round 1 \\ Agent 1\end{tabular}} & Reason & To allocate each person efficiently, we need to consider their strengths and weaknesses. ... %Angela is best at handling customer complaints and queries at the Front Desk due to her calm demeanor, but she is not as efficient with room service tasks and can be disorganized. Miguel excels in room service with his friendly nature but struggles with organization and handling guest queries, which could be challenging at the Front Desk. Vanessa is meticulous, professional, and efficient in both roles but has a stronger background in reception work, which aligns more with Front Desk duties. However, her ability to step in and efficiently complete room service tasks when needed makes her a versatile asset. Considering these aspects, the most efficient allocation would be to utilize Angela's strengths at the Front Desk, where her customer service skills can shine, and pair her with Vanessa, who can manage the front desk efficiently and step in when needed for room service. 
Miguel's charm and friendliness are best utilized in room service, where he can interact with guests in a more relaxed environment and work at his own pace without the strict structure of the front desk. \\ \cline{3-4}
 &  & Answer & Room Service: Miguel, Front Desk: Angela and Vanessa \\ \cline{2-4}
 & \multirow{2}{*}{\begin{tabular}[c]{@{}l@{}}Round 1 \\ Agent 2\end{tabular}} & Reason & To efficiently allocate tasks, we need to consider the strengths and weaknesses of each person. ... %Angela is great at handling customer complaints and queries but can be disorganized and occasionally mislays minor notes. Miguel is charming and good-natured but can be disorganized and has a hard time deciphering guests' queries. Vanessa is meticulous, professional, and friendly but can be overwhelming to Angela. Considering these traits, assigning Angela and Miguel to the Front Desk would not be ideal due to their potential to frustrate each other and Miguel's unfamiliarity with the front desk. Assigning Vanessa to Room Service would utilize her attention to detail and ability to diffuse potential discomfort. Assigning Angela and Miguel to Room Service would also not be ideal due to Angela's occasional disorganization and Miguel's potential to leave tasks incomplete. 
 Therefore, the most efficient allocation would be to assign Vanessa to Room Service, where her attention to detail and professionalism can shine, and Angela and Miguel to the Front Desk, where Angela's customer service skills can be utilized and Miguel's charm can be an asset. \\ \cline{3-4}
 &  & Answer & Room Service: Vanessa, Front Desk: Angela and Miguel \\ \cline{2-4}
 & \multirow{2}{*}{\begin{tabular}[c]{@{}l@{}}Round 1 \\ Agent 3\end{tabular}} & Reason & To allocate each person efficiently, we need to consider their strengths and weaknesses. ... % Angela is great at handling customer complaints and queries but can be disorganized and occasionally mislays minor notes. Miguel is charming and friendly but can be disorganized and has a hard time deciphering guests' queries. Vanessa is meticulous and professional but can be overwhelming and steps on others' toes. Given these traits, Angela's strengths are best utilized at the Front Desk where she can handle customer interactions, while Miguel's charm can be leveraged in Room Service. Vanessa's attention to detail and professionalism make her a strong fit for Room Service as well, where she can ensure tasks are completed efficiently. 
 However, considering the need for balance and the fact that Miguel might struggle with the structure of the Front Desk, it's better to pair Angela with Miguel at the Front Desk, where Angela's organizational skills can help Miguel, and Vanessa can handle Room Service, utilizing her strengths without overwhelming others. \\ \cline{3-4}
 &  & Answer & Room Service: Vanessa, Front Desk: Angela and Miguel \\ \cline{2-4}
 & \multirow{2}{*}{\begin{tabular}[c]{@{}l@{}}Round 2 \\ Agent 1\end{tabular}} & Reason & Considering the strengths and weaknesses of each person, Angela's customer service skills are best utilized at the Front Desk, but her disorganization could be mitigated by pairing her with a detail-oriented person. ... %Miguel's charm is better suited for Room Service, where he can interact with guests in a more relaxed environment, but his disorganization and difficulty with guest queries are concerns. Vanessa's professionalism, attention to detail, and friendliness make her a strong fit for either role, but her tendency to overwhelm others, particularly Angela, suggests she should work independently in Room Service. 
 Given these considerations and the analysis from other agents, it's clear that Angela and Miguel's skills complement each other at the Front Desk, with Angela handling customer interactions and Miguel providing charm, while Vanessa's strengths are best utilized in Room Service, ensuring efficiency and attention to detail. \\ \cline{3-4}
 &  & Answer & Room Service: Vanessa, Front Desk: Angela and Miguel \\ \cline{2-4}
 & \multirow{2}{*}{\begin{tabular}[c]{@{}l@{}}Round 2 \\ Agent 2\end{tabular}} & Reason & Considering the strengths and weaknesses of each person, Angela's customer service skills make her a strong fit for the Front Desk, while Miguel's charm can be leveraged in Room Service. ... %However, Miguel's disorganization and difficulty with guest queries make him a less ideal fit for the Front Desk. Vanessa's attention to detail and professionalism make her a strong fit for either role, but her ability to step in and complete room service tasks efficiently and her background in reception work suggest she could excel at the Front Desk. Given the need to balance the strengths and weaknesses of each team member, assigning Angela and Vanessa to the Front Desk could lead to conflicts due to their differing work styles. Assigning Miguel to Room Service leverages his strengths, but his disorganization could impact efficiency. 
 The most efficient allocation, considering all perspectives, would be to assign Vanessa to Room Service, where her attention to detail can ensure tasks are completed efficiently, and Angela and Miguel to the Front Desk, where Angela's customer service skills and organizational abilities can help mitigate Miguel's weaknesses. \\ \cline{3-4}
 &  & Answer & Room Service: Vanessa, Front Desk: Angela and Miguel \\ \cline{2-4}
 & \multirow{2}{*}{\begin{tabular}[c]{@{}l@{}}Round 2 \\ Agent 3\end{tabular}} & Reason & Considering the strengths and weaknesses of each person, Angela's calm demeanor and customer service skills make her a strong fit for the Front Desk. ... % However, her disorganization and occasional mislaying of minor notes could be mitigated by pairing her with a detail-oriented person. Miguel's charm and friendliness are assets in Room Service, but his disorganization and difficulty with guest queries could be challenging. Vanessa's professionalism, attention to detail, and ability to diffuse potential discomfort make her versatile and able to excel in either role. Given the analysis from other agents, it's clear that Angela and Miguel at the Front Desk could be problematic due to their potential to frustrate each other and Miguel's unfamiliarity with the front desk. 
 Assigning Vanessa to Room Service leverages her strengths, and pairing Angela with Miguel at the Front Desk utilizes Angela's customer service skills and Miguel's charm, with Angela's organizational skills potentially helping to mitigate Miguel's disorganization. \\ \cline{3-4}
 &  & Answer & Room Service: Vanessa, Front Desk: Angela and Miguel \\ \cline{2-4}
& \multicolumn{2}{>{\centering\arraybackslash}p{0.2\textwidth}|}{\textbf{Final Answer}}
 %& & \textbf{Final Answer} 
 & Room Service: Vanessa, Front Desk: Angela and Miguel \\ \midrule[1.5pt]
\multirow{3}{*}{DOWN} & \multirow{2}{*}{\begin{tabular}[c]{@{}l@{}}Round 1 \\ Agent 1\end{tabular}} & Reason & To efficiently allocate tasks, it's essential to consider the strengths and weaknesses of each team member. Angela excels in customer service, particularly at the Front Desk, due to her calm demeanor and respectful tone. ... % However, her occasional misplacement of minor notes and mixed reviews for room service suggest she might not be the best fit for Room Service. Miguel's charm and ability to work at his own pace make him suitable for Room Service, where he can interact with guests in a more relaxed environment. Nevertheless, his difficulty in deciphering guests' queries and concerns, as well as his unfamiliarity with the Front Desk, indicate that he should not be assigned to the Front Desk. Vanessa's meticulous attention to detail, professionalism, and experience at a high-end hotel reception make her an ideal candidate for the Front Desk. Her ability to diffuse potential discomfort and balance professionalism with friendliness is crucial for guest interactions at the Front Desk. 
Additionally, her efficiency and thoroughness in room service tasks, although sometimes stepping on Angela's toes, demonstrate her capability to handle detailed work. \\ \cline{3-4}
 &  & Answer & Room Service: Miguel, Front Desk: Angela and Vanessa \\ \cline{3-4}
 &  & Confidence Score & 0.91 \\ \cline{2-4}
& \multicolumn{2}{>{\centering\arraybackslash}p{0.2\textwidth}|}{\textbf{Final Answer}}
  & Room Service: Miguel, Front Desk: Angela and Vanessa
\\\bottomrule[1.5pt]
\caption{Qualitative analysis of the MUSR dataset}
    \label{tb:qual3}
\end{longtable}
%\begin{table}[h]
%    \caption{Qualitative analysis of the MUSR dataset}
%    \label{tb:qual3}
%\end{table}
}

\end{document}